\documentclass{bmvc2k}

\newcommand{\final}{1}

\usepackage{graphicx}
\usepackage{float}
\usepackage[justification=raggedright]{caption}	
\usepackage{lscape}                                         

\usepackage[lined,ruled,linesnumbered]{algorithm2e}

\usepackage{booktabs}                   
\usepackage{multirow}

\usepackage{paralist}
\usepackage{enumitem}

\usepackage{bm}                          
\usepackage{epsfig}                      
\usepackage{graphicx}                  
\usepackage{times}
\usepackage{mathptmx}
\usepackage{mathtools}
\usepackage{amssymb,amsmath}   

\usepackage{units}
\usepackage{color}

\usepackage{comment}

\usepackage{url}  
\usepackage{xspace}
\usepackage[table]{xcolor}
\usepackage{setspace}

\def\etal{et~al.}			  
\def\eg{e.g.,~}               
\def\ie{i.e.,~}               
\def\vs{vs.~}                 

\newlength\paramargin
\newlength\figmargin
\newlength\secmargin
\newlength\figcapmargin
\newlength\tabmargin
\newlength\figbmargin
\newlength\figwidth
\newcommand{\mpage}[2]
{
\begin{minipage}{#1\linewidth}\centering
#2
\end{minipage}
}

\newcommand {\para}[1]{\vspace{1.5mm} \noindent \textbf{#1}}

\newcommand{\secref}[1]{Section~\ref{sec:#1}}
\newcommand{\figref}[1]{Figure~\ref{fig:#1}} 
\newcommand{\tabref}[1]{Table~\ref{tab:#1}}

\long\def\ignorethis#1{}
\newcommand {\jiabin}[1]{{\color{blue}\textbf{Jia-Bin: }#1}\normalfont}
\newcommand {\chen}[1]{{\color{red}\textbf{Chen: }#1}\normalfont}
\newcommand {\yl}[1]{{\color{magenta}\textbf{Yuliang: }#1}\normalfont}

\ifthenelse{\equal{\final}{1}}
{
\renewcommand{\jiabin}[1]{}
\renewcommand{\chen}[1]{}
\renewcommand{\yl}[1]{}
}
{}

\newcommand{\tb}[1]{\textbf{#1}}

\def\xi{\mathbf{x}_i}

\graphicspath{{figure}, {example}}

\usepackage[utf8]{inputenc}
\usepackage{listings}
 \definecolor{codegreen}{rgb}{0,0.6,0}
\definecolor{codegray}{rgb}{0.5,0.5,0.5}
\definecolor{codepurple}{rgb}{0.58,0,0.82}
\definecolor{backcolour}{rgb}{0.95,0.95,0.92}
 \lstdefinestyle{mystyle}{
    backgroundcolor=\color{backcolour},   
    commentstyle=\color{codegreen},
    keywordstyle=\color{magenta},
    numberstyle=\tiny\color{codegray},
    stringstyle=\color{codepurple},
    basicstyle=\footnotesize,
    breakatwhitespace=false,         
    breaklines=true,                 
    captionpos=b,                    
    keepspaces=true,                 
    numbersep=5pt,                  
    showspaces=false,                
    showstringspaces=false,
    showtabs=false,                  
    tabsize=1
}
\title{iCAN: Instance-Centric Attention Network\\ for Human-Object Interaction Detection}

\addauthor{Chen Gao}{chengao@vt.edu}{1}
\addauthor{Yuliang Zou}{ylzou@vt.edu}{1}
\addauthor{Jia-Bin Huang}{jbhuang@vt.edu}{1}

\addinstitution{
 Virginia Tech\\
 Virginia, USA
}

\runninghead{Gao~\etal}{Instance-Centric Attention Network}

\begin{document}

\maketitle
\begin{center}
\centering
\mpage{0.31}{\includegraphics[width=\linewidth]{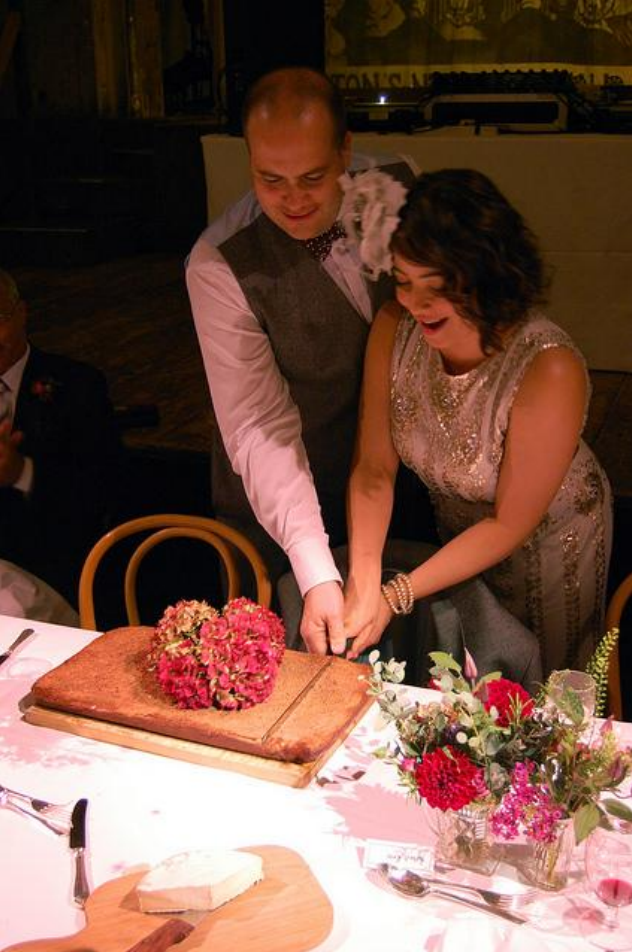}}\hfill
\mpage{0.31}{\includegraphics[width=\linewidth]{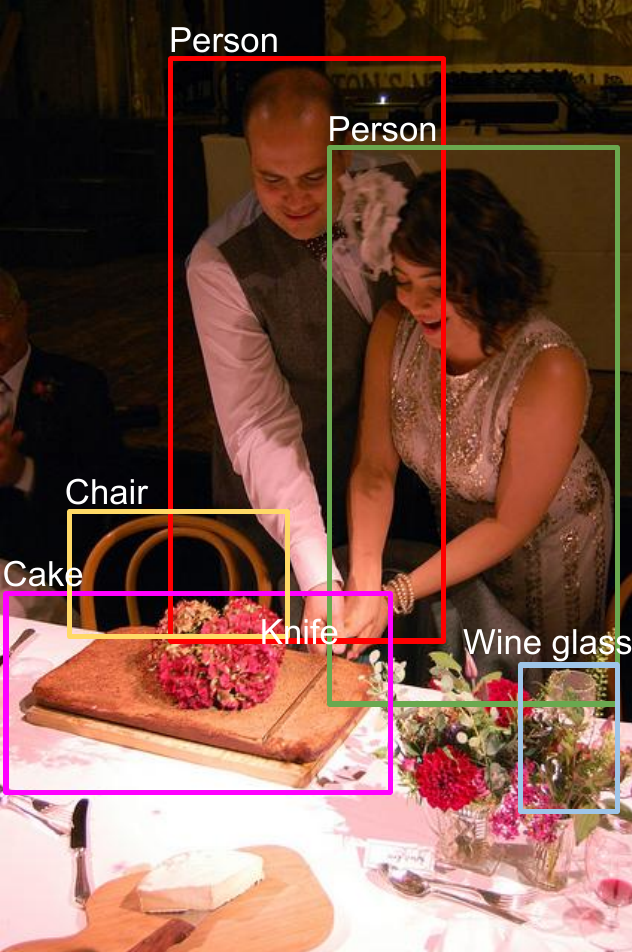}}\hfill
\mpage{0.31}{\includegraphics[width=\linewidth]{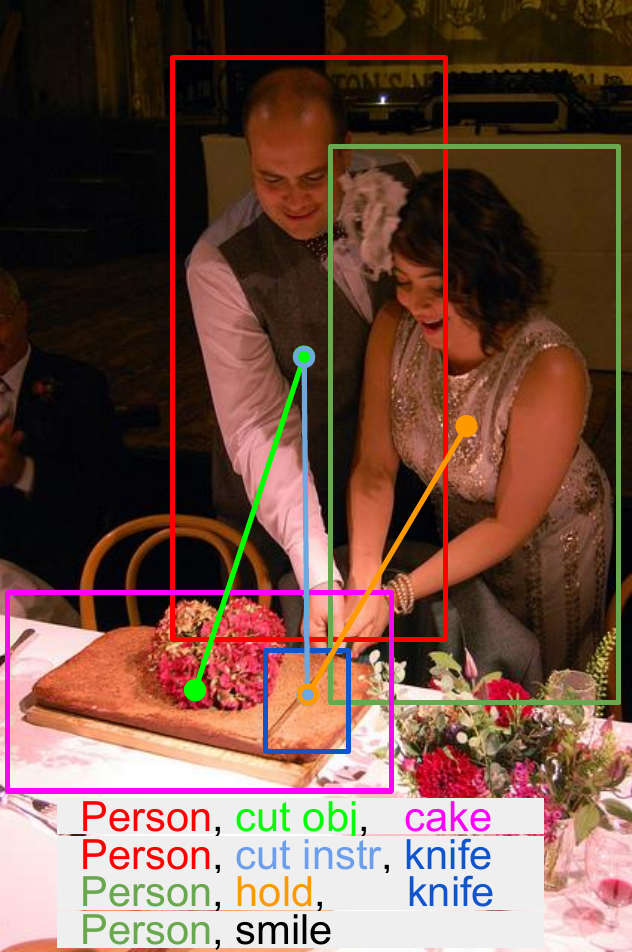}} \\
\vspace{1mm}
\mpage{0.31}{Input image} \hfill
\mpage{0.31}{Object detection} \hfill
\mpage{0.31}{HOI detection} \\ 
\vspace{\figmargin}
\captionof{figure}{
\textbf{Human-object interaction detection.} Given an input image (\emph{left}) and the detected object instances in the image (\emph{middle}), our method detects and recognizes the interactions between each person and the objects they are interacting with (\emph{right}). 
}
\label{fig:teaser}
\end{center}

\begin{abstract}
Recent years have witnessed rapid progress in detecting and recognizing individual object instances.
To understand the situation in a scene, however, computers need to recognize how humans interact with surrounding objects.
In this paper, we tackle the challenging task of detecting human-object interactions (HOI).
Our core idea is that the appearance of a person or an object instance contains informative cues on which relevant parts of an image to attend to for facilitating interaction prediction.
To exploit these cues, we propose an instance-centric attention module that learns to dynamically highlight regions in an image conditioned on the appearance of each instance.
Such an attention-based network allows us to selectively aggregate features relevant for recognizing HOIs.
We validate the efficacy of the proposed network on the Verb in COCO and HICO-DET datasets and show that our approach compares favorably with the state-of-the-arts.
\end{abstract}

\section{Introduction}
\label{sec:intro}

Over the past few years, there has been rapid progress in visual recognition tasks, including object detection~\cite{Dai-NIPS-RFCN,Ross-CVPR-FastRCNN,Lin-CVPR-Pyramid,Ren-NIPS-FasterRCNN}, segmentation~\cite{Chen-TPAMI-DeepLab,Ross-CVPR-Hierarchies,He-ICCV-MaskRCNN,Long-CVPR-FCN}, and action recognition~\cite{Cheron-ICCV-PCNN,Girdhar-NIPS-AttentionalPooling,Gkioxari-ICCV-R*CNN,Maji-CVPR-Action,Yao-CVPR-Combining}.
However, understanding a scene requires not only detecting individual object instances but also recognizing the visual relationship between object pairs.
One particularly important class of visual relationship detection is detecting and recognizing how each person interacts with the surrounding objects.
This task, known as Human-Object Interactions (HOI) detection~\cite{Chao-WACV-HOI,Gkioxari-CVPR-InteractNet,Gupta-TPAMI-Observing,Gupta-SemanticRoleLabeling}, aims to localize a person, an object, as well as identify the interaction between the person and the object.
%
%
In~\figref{teaser}, we show an example of the HOI detection problem.
Given an input image and the detected instances from an object detector, we aim to identify all the triplets $\langle$ $\texttt{human, verb, object}$ $\rangle$.
%

\textbf{Why HOI?} Detecting and recognizing HOI is an essential step towards a deeper understanding of the scene.
Instead of ``What is where?'' (\ie localizing object instances in an image), the goal of HOI detection is to answer the question ``What is happening?''.
Studying the HOI detection problem also provides important cues for other related high-level vision tasks, such as pose estimation~\cite{Cao-CVPR-OpenPese,Yao-BMVC-Pose}, image captioning~\cite{Li-CVPR-Scene,Xu-CVPR-SceneGraph}, and image retrieval~\cite{Johnson-CVPR-Retrieval}.
%

\begin{figure}[t]
\centering

\mpage{0.33}{\includegraphics[width=\linewidth]{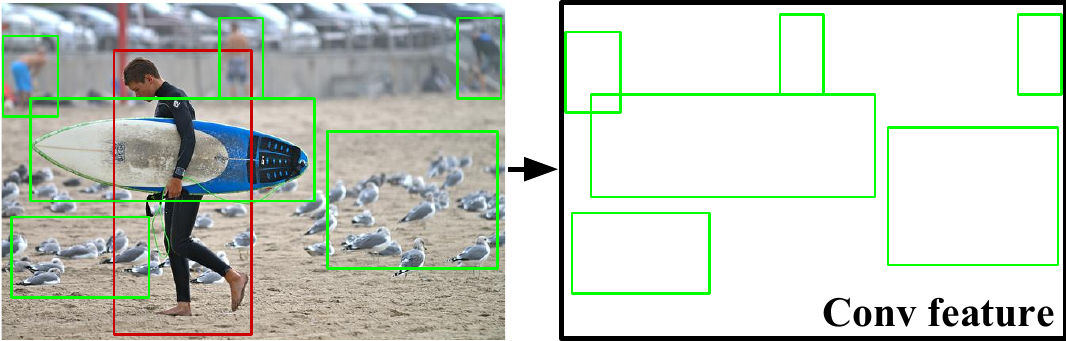}}
\mpage{0.65}{\includegraphics[width=\linewidth]{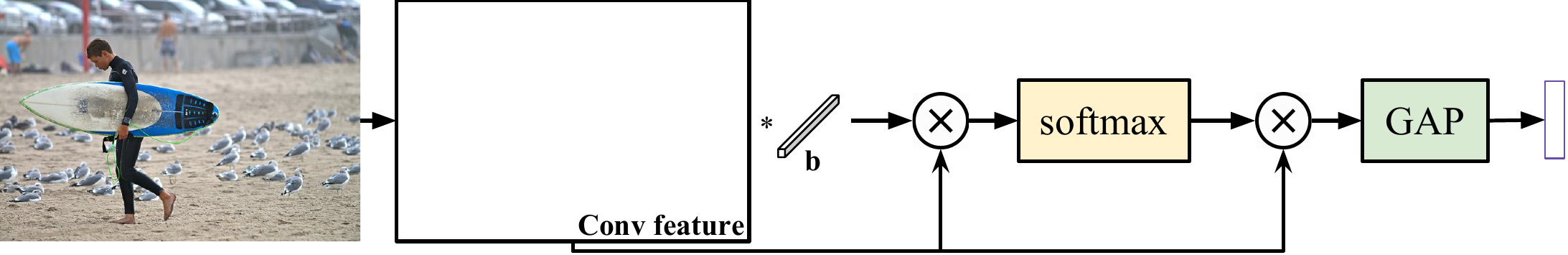}}\hfill

\mpage{0.33}{Secondary regions~\cite{Gkioxari-ICCV-R*CNN}}\hfill
\mpage{0.6}{Bottom-up attentional feature~\cite{Girdhar-NIPS-AttentionalPooling}}\hfill



\mpage{0.33}{\includegraphics[width=\linewidth]{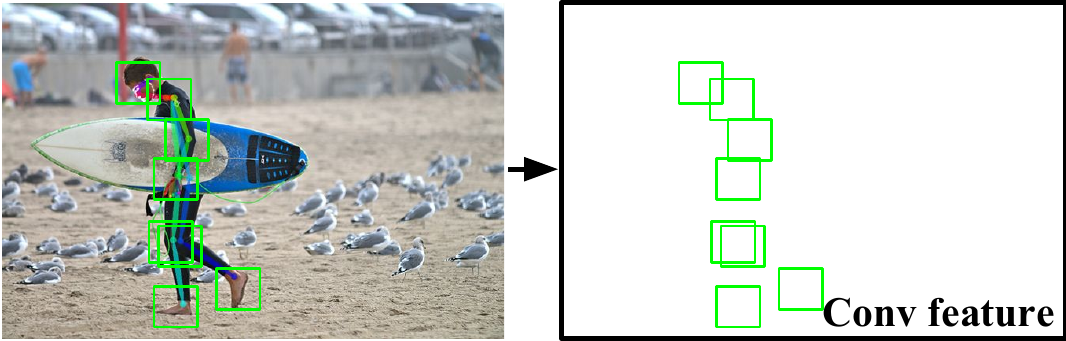}}
\mpage{0.65}{\includegraphics[width=\linewidth]{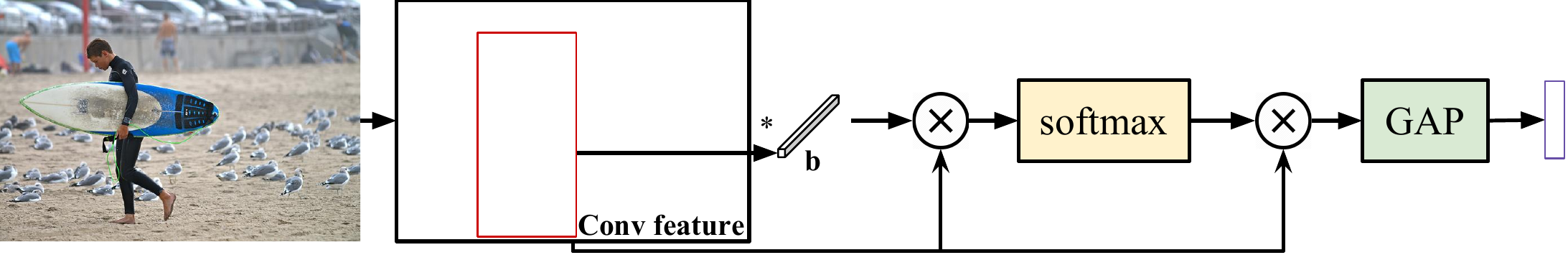}}\hfill

\mpage{0.35}{Human pose~\cite{Cheron-ICCV-PCNN}}\hfill
\mpage{0.55}{Instance-centric attentional feature (ours)}\hfill
\\
\vspace{\figmargin}
\captionof{figure}{\textbf{Examples of contextual features.} Different ways of capturing contextual cues from an image in addition to using the bounding boxes of persons and objects.
}
\vspace{\figbmargin}
\label{fig:scene}
\end{figure}

\textbf{Why attention?} 
Driven by the progress in object detection~\cite{He-ICCV-MaskRCNN,Ren-NIPS-FasterRCNN}, several recent efforts have been devoted to detecting HOI in images~\cite{Chao-WACV-HOI,Gkioxari-CVPR-InteractNet,Gupta-SemanticRoleLabeling,Shen-WACV-Zeroshot}.
Most existing approaches infer interactions using appearance features of a person and an object as well as their spatial relationship.
In addition to using only appearance features from a person, recent action recognition algorithms exploit contextual cues from an image.
As shown in~\figref{scene}, examples of encoding context include selecting a secondary box~\cite{Gkioxari-ICCV-R*CNN}, using the union of the human and object bounding boxes~\cite{Lu-ECCV-Prior}, extracting features around human pose keypoints~\cite{Cheron-ICCV-PCNN}, or exploiting global context from the whole image~\cite{Mallya-ECCV-Interactions}.
While incorporating context generally helps improve performance, these hand-designed attention regions may not always be relevant for recognizing actions/interactions.
For examples, attending to human poses may help identify actions like `ride' and `throw', attending to the point of interaction may help recognize actions involving hand-object interaction such as `drinking with cup' and `eat with spoon', and attending to the background may help distinguish between `hit with tennis racket' and `hit with baseball ball bat'.
%
To address this limitation, recent works leverage end-to-end trainable attention modules for action recognition~\cite{Girdhar-NIPS-AttentionalPooling} or image classification~\cite{Jetley-ICLR-PayAttention}.
These methods, however, are designed for \emph{image-level} classification tasks.
%
%
%

\textbf{Our work.} In this paper, we propose an end-to-end trainable \emph{instance-centric} attention module that learns to highlight informative regions using the appearance of a person or an object instance.
Our intuition is that the appearance of an instance (either human or an object) provides cues on where in the image we should pay attention to.
For example, to better determine whether a person is carrying an object, one should direct its attention to the region around the person's hands.
On the other hand, given a bicycle in an image, attending to the pose of the person nearby helps to disambiguate the potential interactions involved with object instance (\eg riding or carrying a bike).
%
The proposed instance-centric attention network (iCAN) dynamically produces an attentional map for each detected person or object instance highlighting regions relevant to the task.
%
%
We validate the efficacy of our network design on two large public benchmarks on HOI detection: Verbs in COCO (V-COCO)~\cite{Gupta-SemanticRoleLabeling} and Humans Interacting with Common Objects (HICO-DET)~\cite{Chao-WACV-HOI} datasets.
\jiabin{HICO results need to be updated. I also updated the V-COCO improvement because we added BAR-CNN.}
Our results show that the proposed iCAN compares favorably against the state-of-the-art with around 10\% relative improvement on V-COCO and 49\% on HICO-DET with respect to the existing best-performing methods.

\textbf{Our contributions.} We make the following four contributions.
\begin{compactitem}
\item We introduce an instance-centric attention module that allows the network to dynamically highlight informative regions for improving HOI detection.
%
\item We establish new state-of-the-art performance on two large-scale HOI benchmark datasets.
%
\item We conduct detailed ablation study and error analysis to identify the relative contributions of the individual components and quantify different types of errors. 
\item We release our source code and pre-trained models to facilitate future research. 
\footnote{Project webpage: \url{https://gaochen315.github.io/iCAN/}}
\end{compactitem}

\vspace{\secmargin}
\section{Related Work}
\label{sec:related}
\vspace{\secmargin}

%

\para{Object detection.}
%
Object detection~\cite{Dai-NIPS-RFCN,Ross-CVPR-FastRCNN,Ross-CVPR-Hierarchies,Lin-CVPR-Pyramid,Ren-NIPS-FasterRCNN} is an essential building block for scene understanding.
Our work uses the off-the-shelf Faster R-CNN~\cite{Ren-NIPS-FasterRCNN,Detectron} to localize persons and object instances. 
Given the detected instances, our method aims to recognize interactions (if any) between all pairs of person and object instances.
%
%

\para{Visual relationship detection.} 
A number of recent work addresses the problem of detecting visual relationship ~\cite{Bilen-CVPR-Weakly,Dai-CVPR-Relationship,Hu-CVPR-Referential,Li-CVPR-VIP,Zhang-ICCV-PPR,Peyre-ICCV-Weakly,Zhuang-ICCV-ContextAware,kolesnikov2018detecting} and generating scene graph~\cite{Xu-CVPR-SceneGraph,li2017scene,zellers2018neural}.
Several papers leverage some forms of language prior~\cite{Lu-ECCV-Prior,Plummer-ICCV-Phrase} to help overcome the problem of large numbers of the relationship subject-predicate-object triplets and limited data samples.
%
Our work focuses on one particular class of visual relationship detection problems: detecting human-object interactions.
%
HOI detection poses additional challenges over visual relationship detection. 
With human as a subject, the interactions (\ie the predicate) with objects are a lot more fine-grained and diverse than other generic objects.

\para{Attention.} 
Extensive efforts have been made to incorporate attention in action recognition~\cite{Gkioxari-ICCV-R*CNN,Cheron-ICCV-PCNN} and human-object interaction tasks~\cite{Lu-ECCV-Prior,Mallya-ECCV-Interactions}.
These methods often use hand-designed attention regions to extract contextual features. 
Very recently, end-to-end trainable attention-based methods have been proposed to improve the performance of action recognition~\cite{Girdhar-NIPS-AttentionalPooling} or image classification~\cite{Jetley-ICLR-PayAttention}.
However, these methods are designed for \textit{image-level} classification task.
Our work builds upon the recent advances of attention-based techniques and extends them to address \emph{instance-level} HOI recognition tasks.
%
%
%
%

\para{Human-object interactions.}
Detecting HOI provides a deeper understanding of the situation in a scene. 
%
Gupta and Malik~\cite{Gupta-SemanticRoleLabeling} first tackle the HOI detection problem --- detecting people doing actions and the object instances they are interacting with.
%
Associating objects in a scene with various semantic roles leads to a finer-grained understanding of the current state of activity. 
Very recently, Gkioxari~\etal~\cite{Gkioxari-CVPR-InteractNet} extend the method in \cite{Gupta-SemanticRoleLabeling} by introducing an action-specific density map over target object locations based on the appearance of a detected person.
Significantly improved results have also been shown by replacing feature backbone with ResNet-50~\cite{He-CVPR-ResNet} and the Feature Pyramid Network~\cite{Lin-CVPR-Pyramid}.
In addition to using object instance appearances, Chao~\etal~\cite{Chao-WACV-HOI} also encode the relative spatial relationship between a person and the object with a CNN.
%
Our work builds upon these recent advances in HOI detection, but with a key differentiator.
Existing work recognizes interactions based on individual cues (either human appearance, object appearance, or spatial relationship between a human-object pair).
Our key observation is that such predictions inevitably suffer from the lack of contextual information.
The proposed instance-centric attention module extracts contextual features complementary to the appearance features of the localized regions (\eg humans/object boxes) to facilitate HOI detection.


\vspace{\secmargin}
\section{Instance-Centric Attention Network}
\label{sec:approach}
\vspace{\secmargin}

\begin{figure*}[t]
\centering
\includegraphics[width=1.0 \linewidth]{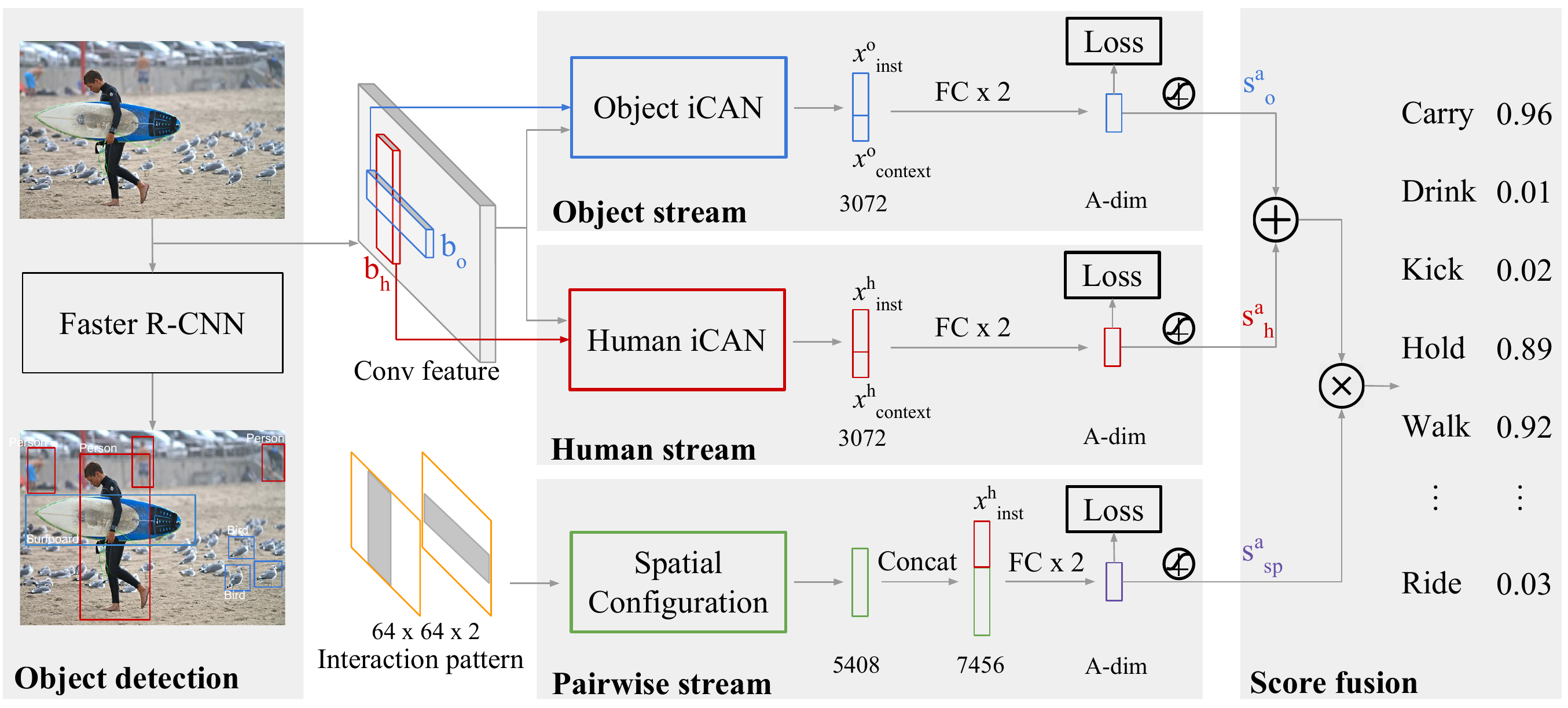}

\vspace{\figmargin}
\caption{\tb{Overview of the proposed model.} 
The proposed model consists of following three major streams:
%
%
%
(1) a \textit{human stream} for detecting interaction based on human appearance;
(2) an \textit{object steam} that predicts the interaction based on object appearance;
(3) a \textit{pairwise stream} for encoding the spatial layouts between the human and object bounding boxes.
Given the detected object instances by the off-the-shelf Faster R-CNN, we generate the HOI hypothesis using all the human-object pairs.
The action scores from individual streams are then fused to produce the final prediction as shown on the right.
}
\vspace{\figbmargin}
\label{fig:overview}
\end{figure*}

In this section, we present our Instance-centric Attention Network for HOI detection (\figref{overview}).
%
%
We start with an overview of our approach (\secref{overview}) and then introduce the instance-centric attention module (\secref{attention}). 
Next, we outline the details of the three main streams for feature extraction (\secref{multi-stream}): the human stream, the object stream, and the pairwise stream.
Finally, we describe our inference procedure (\secref{inference}).

\begin{figure*}[t]
\centering
\includegraphics[width=1.0 \linewidth]{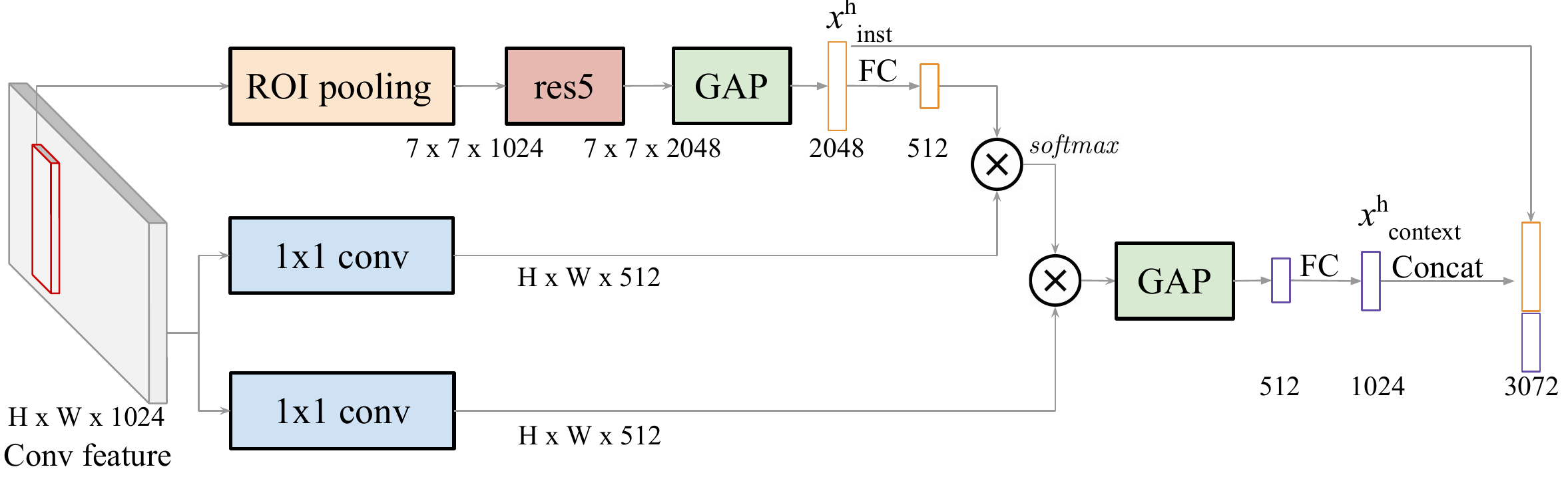}

\vspace{\figmargin}
\caption{\tb{iCAN module.} Given the convolutional features of the image (shown in gray) and a human/object bounding box (shown in red), the iCAN module extracts the appearance features of the instance $x^\mathrm{h}_\mathrm{inst}$ (for human) or $x^\mathrm{o}_\mathrm{inst}$ (for object) as well as the features from the instance-centric attentional map.
For computing the attentional map, we measure the similarity in the embedding space with a bottleneck of 512 channels~\cite{vaswani2017attention,wang2017non}.
Specifically, we embed the image feature using a $1 \times 1$ convolution and the instance appearance feature $x^\mathrm{h}_\mathrm{inst}$ with a fully connected layer.
%
%
Here \texttt{res5} denotes the fifth residual block, \texttt{GAP} denotes a global average pooling layer, and \texttt{FC} denotes a fully connected layer. 
%
%
}
\vspace{\figbmargin}
\label{fig:iCAN}
\end{figure*}

\vspace{\secmargin}
\subsection{Algorithm overview}
\label{sec:overview}
\vspace{\secmargin}

Our approach to human-object interaction detection consists of two main steps: 1) object detection and 2) HOI prediction.
%
First, given an input image we use Faster R-CNN~\cite{Ren-NIPS-FasterRCNN} from Detectron~\cite{Detectron} to detect all the person/object instances.
We denote $b_h$ as the detected bounding box for a person and $b_o$ for an object instance.
We use $s_h$ and $s_o$ to denote the confidence scores for a detected person and an object, respectively.
Second, we evaluate all the human-object bounding box pairs through the proposed instance-centric attention network to predict the interaction score.
\figref{overview} shows an overview of the model.

\para{Inference.} 
We predict HOI scores in a similar manner to the existing methods~\cite{Gupta-SemanticRoleLabeling,Gkioxari-CVPR-InteractNet}.
%
%
For each human-object bounding box pair ($b_h$, $b_o$), we predict the score $S^a_{h,o}$ for each action $a \in \left\{ 1, \cdots, A \right\}$, where $A$ denotes the total number of possible actions.
The score $S^a_{h,o}$ depends on  
(1) the confidence for the individual object detections ($s_h$ and $s_o$), 
(2) the interaction prediction based on the appearance of the person $s^a_h$ and the object $s^a_h$, and 
(3) the score prediction based on the spatial relationship between the person and the object $s_{sp}^a$.
Specifically, our HOI score $S^a_{h,o}$ for the human-object bounding box pair ($b_h$, $b_o$) has the form:
\begin{equation}
S^a_{h,o} = s_h \cdot s_o \cdot (s^a_{h} + s^a_{o}) \cdot s^a_{sp}.
\end{equation}
For some of the action classes that do not involve any objects (\eg walk, smile), we use the action score $s^a_h$ from the human stream only. 
For those actions, our final scores are $s_h \cdot s^a_h$.

\para{Training.} 
As a person can concurrently perform different actions to one or multiple target objects, \eg a person can 'hit with' and 'hold' a tennis racket at the same time, HOI detection is thus a \emph{multi-label classification} problem, where each interaction class is independent and not mutually exclusive. 
We apply a binary sigmoid classifier for each action category, and then minimize the cross-entropy loss between action score $s^a_{h}$, $s^a_{o}$, or $s^a_{sp}$ and the ground-truth action label \emph{for each action category}.
%
In the following, we introduce our instance-centric attention module for extracting informative features from an image and then describe a multi-stream network architecture for computing the action scores $s^a_{h}$, $s^a_{o}$, and $s^a_{sp}$.

%
%
%
%
\jiabin{Same here. Pairwise stream is not defined yet.}
\subsection{Instance-centric attention module}
\label{sec:attention}
\vspace{\secmargin}


In this section, we introduce the instance-centric attention module for extracting contextual features from an image.
\figref{iCAN} shows the detailed procedure using human as an instance for clarity.
Using an object as an instance is straightforward.

We first extract the instance-level appearance feature $x^h_\mathrm{inst}$ using the standard process, e.g., applying ROI pooling, passing through a residual block, followed by the global average pooling. 
Next, our goal is to dynamically generate an attention map conditioned on the object instance of interest. 
To do so, we embed both the instance-level appearance feature $x^h_\mathrm{inst}$ and the convolutional feature map onto a 512-dimensional space and measure the similarity in this embedding space using vector dot product.
We can then obtain the instance-centric attentional map by applying softmax. 
The attentional map highlights relevant regions in an image that may be helpful for recognizing HOI associated with the given human/object instance.
Using the attentional map, we can extract the contextual feature $x^h_\mathrm{context}$ by computing the weighted average of the convolutional features.
The final output of our iCAN module is a concatenation of instance-level appearance feature $x^h_\mathrm{inst}$ and the attention-based contextual feature $x^h_\mathrm{context}$.

Our iCAN module offers several advantages over existing approaches.
First, unlike hand-designed contextual features based on pose~\cite{Cheron-ICCV-PCNN}, the entire image~\cite{Mallya-ECCV-Interactions}, or secondary regions~\cite{Gkioxari-ICCV-R*CNN}, our attention map are automatically learned and jointly trained with the rest of the networks for improving the performance.
Second, when compared with attention modules designed for image-level classification, our \emph{instance-centric} attention map provides greater flexibility as it allows attending to different regions in an image depending on different object instances.
\vspace{\secmargin}
\subsection{Multi-stream network}
\label{sec:multi-stream}
\vspace{\secmargin}


As shown in \figref{overview}, our network uses three streams to compute the action scores based on human appearance $s^a_h$, object appearance $s^a_o$, and their spatial relationship $s^a_{sp}$.

\para{Human/object stream.} For human and object stream,  we extract both 
1) the instance-level appearance feature $x^h_\mathrm{inst}$ for a person or $x^o_\mathrm{inst}$ for an object and 
2) the contextual features $x^h_\mathrm{context}$ (or $x^o_\mathrm{context}$) based on the  attentional map following the steps outlined in \secref{attention} and \figref{iCAN}.
%
%
%
With the two feature vectors, we then concatenate them and pass it through two fully connected layers to produce the action scores $s^a_{h}$ and $s^a_{o}$.
The score $s^a_{h}$ from the human stream also allows us to detect actions that do not involve any objects, e.g., walk, smile.

%


\para{Pairwise stream.}
%
While the human and object appearance features contain strong cues for recognizing the interaction, using appearance features alone often leads to plausible but incorrect predictions.
%
%
%
To encode the spatial relationship between the person and object, we adopt the two-channel binary image representation in~\cite{Chao-WACV-HOI} to characterize the interaction patterns. 
Specifically, we take the union of these two boxes as the reference box and construct a binary image with two channels within it. 
%
%
The first channel has value 1 within the human bounding box and value 0 elsewhere; the second channel has value 1 within the object bounding box and value 0 elsewhere. 
%
We then use a CNN to extract spatial features from this two-channel binary image.
However, we found that this feature by itself cannot produce accurate action prediction due to the coarse spatial information (only two bounding boxes).
To address this, we concatenate the spatial feature with the human appearance feature $x^h_\mathrm{inst}$.
Our intuition is that the appearance of the person can greatly help disambiguate different actions with similar spatial layouts, e.g., riding \vs walking a bicycle.
%
%

\subsection{Efficient inference}
\label{sec:inference}
\vspace{\secmargin}

%
Following Gkioxari \etal~\cite{Gkioxari-CVPR-InteractNet} we compute the scores for the triplets in a cascade fashion.
%
We first compute the scores from the human and the object stream action classification head, for each box $b_h$ and $b_o$, respectively. 
This first step has a complexity of $O(n)$ for $n$ human/object instances.
The second step involves computing scores of all possible human-object pairs.
While the second step has a complexity of $O(n^2)$, computing the scores $S^a_{h,o}$, however, is very efficient as it involves summing a pair of scores from the human stream $s^a_h$ and object stream $s^a_o$ (which are already computed and cached in the first step).

\para{Late \vs early fusion.} We refer to our approach using the pairwise summing scores method as \emph{late fusion} (because the action scores are independently predicted from the human/object streams first and then summed later).
We also implement a variant of iCAN with \emph{early fusion}. 
Specifically, we first concatenate all the features from human iCAN, object iCAN, and the pairwise stream and use two fully connected layers to predict the action score.
Unlike late fusion, the early fusion approach needs to evaluate the scores from all human-object pairs and thus have slower inference speed and does not scale well for scenes with many objects.

\vspace{\secmargin}
\section{Experimental Results}
\label{sec:results}
\vspace{\secmargin}

We evaluate the performance of our proposed iCAN model and compare with the state-of-the-art on two large-scale HOI benchmark datasets.
Additional results including detailed class-wise performance and error diagnosis can be found in the supplementary material.
The source code and the pre-trained models are available on our project page.
%

\vspace{\secmargin}
\subsection{Experimental setup}
\vspace{\secmargin}

\para{Datasets.}
{V-COCO}~\cite{Gupta-SemanticRoleLabeling} is a subset of the COCO dataset~\cite{Lin-ECCV-MSCOCO} that provides HOI annotations. 
V-COCO includes a total of 10,346 images containing 16,199 human instances. 
Each person is annotated with a binary label vector for 26 different actions (where each entry indicates whether the person is performing a certain action).
Each person can perform multiple actions at the same time, e.g., holding a cup while sitting on a chair.
%
%
\ignorethis{
Three action classes (\textit{cut}, \textit{hit}, \textit{eat}) are annotated with two types of target objects: (1) instrument and (2) direct object. 
For example, \textit{eat} + \textit{fork} involves the instrument (meaning ``eat with a fork"), and \textit{eat} + \textit{pizza} involves the direct object (meaning `eat a pizza').
To address this issue, we consider these three actions with different targets as independent actions, resulting in a total of 29 action classes.
}
%
{HICO-DET}~~\cite{Chao-CVPR-HICO} is a subset of the HICO dataset~\cite{Chao-CVPR-HICO}.
HICO-DET contains 600 HOI categories over 80 object categories (same as \cite{Lin-ECCV-MSCOCO}), and provides more than 150K annotated instances of human-object pairs.

\para{Evaluation metrics.}
We evaluate the HOI detection performance using the commonly used role mean average precision (role mAP)~~\cite{Gupta-SemanticRoleLabeling} for both V-COCO and HICO datasets.
%
%
The goal is to detect the agent and the objects in the various roles for the action, denoted as the $\langle$ $\texttt{human, verb, object}$ $\rangle$ triplet. 
%
%
A detected triplet is considered as a true positive if it has the correct action label, and both the predicted human and object bounding boxes $b_h$ and $b_o$ have IoUs $\geq 0.5$ w.r.t the ground truth annotations.

%

\para{Implementation details.} 
We use Detectron~\cite{Detectron} with a feature backbone of ResNet-50-FPN~\cite{Lin-CVPR-Pyramid} to generate human and object bounding boxes.
We keep human boxes with scores $s_h$ higher than 0.8 and object boxes with scores $s_o$ higher than 0.4.
%
%
We implement our network based on Faster R-CNN~\cite{Ren-NIPS-FasterRCNN} with a ResNet-50~\cite{He-CVPR-ResNet} feature backbone.
\footnote{We believe that using ResNet-50-FPN based on the Detectron framework for jointly training object detection and HOI detection could lead to improved performance.}
%
%
We train our network for 300K iterations on the V-COCO \emph{trainval} set with a learning rate of 0.001, a weight decay of 0.0001, and a momentum of 0.9. 
Training our network on V-COCO takes $16$ hours on a single NVIDIA P100 GPU. 
\jiabin{Please verify the details.}
For HICO-DET, training the network on the \emph{train} set takes $72$ hours.
Using a single NVIDIA P100 GPU, our method (with late score fusion) takes less than 75ms to process an image of size $480 \times 640$ (including ResNet-50 feature extraction, multi-stream network, attention-based feature extraction, and HOI recognition).
%
We apply the \emph{same} training and inference procedures for both V-COCO and HICO-DET datasets.
%
%
Please refer to the supplementary material for additional implementation details.


\vspace{\secmargin}
\subsection{Quantitative evaluation
}
\vspace{\secmargin}

 \begin{table}[t]
\centering
\caption{Performance comparison with the state-of-the-arts on the V-COCO $\emph{test}$ set.}
\label{tab:vcoco_comparison}

\vspace{2mm}
\begin{tabular}{ll|c}
\toprule
Method & Feature backbone & $AP_{role}$ \\
\midrule
Model C of~\cite{Gupta-SemanticRoleLabeling} (implemented by~\cite{Gkioxari-CVPR-InteractNet}) & ResNet-50-FPN & 31.8\\
InteractNet~\cite{Gkioxari-CVPR-InteractNet}      & ResNet-50-FPN & 40.0 \\
BAR-CNN~\cite{kolesnikov2018detecting} & Inception-ResNet~\cite{szegedy2016rethinking}&  41.1 \\
iCAN (ours) w/ late fusion               & ResNet-50  & \underline{44.7} \\
iCAN (ours) w/ early fusion              & ResNet-50  & {\textbf{45.3}} \\
\bottomrule
\end{tabular}
\vspace{\tabmargin}
\end{table}

\begin{table}[t]
\centering
\caption{Performance comparison with the state-of-the-arts on HICO-DET \emph{test} set. The results from our model are from \emph{late fusion}. 
}
\label{tab:HICO}

\vspace{2mm}
\resizebox{\columnwidth}{!}{
\begin{tabular}{l l ccc c  ccc}
\toprule
\multirow{2}{*}{} & &
\multicolumn{3}{c}{Default} & &
\multicolumn{3}{c}{Known Object}\\
\cline{3-5} \cline{7-9}
Method & Feature backbone & Full & Rare & Non Rare &  & Full & Rare & Non Rare  \\
\midrule
Shen~\etal ~\cite{Shen-WACV-Zeroshot} & VGG-19 & 6.46 & 4.24 & 7.12 & & - & - & -\\
HO-RCNN ~\cite{Chao-WACV-HOI} & CaffeNet & 7.81 & 5.37 & 8.54 & & $\underline{10.41}$ & $\underline{8.94}$ & $\underline{10.85}$\\
InteractNet ~\cite{Gkioxari-CVPR-InteractNet} & ResNet-50-FPN & 9.94 & $\underline{7.16}$ & 10.77 & & - & - & -\\
iCAN (ours) & ResNet-50 & $\textbf{14.84}$ & $\textbf{10.45}$ & $\textbf{16.15}$ & & $\textbf{16.26}$ & $\textbf{11.33}$ & $\textbf{17.73}$ \\
\bottomrule
\end{tabular}
}
\vspace{\tabmargin}
\end{table}

We present the overall quantitative results in terms of $AP_{role}$ on V-COCO in \tabref{vcoco_comparison} and HICO-DET in \tabref{HICO}.
For V-COCO, the proposed instance-centric attention network achieves sizable performance boost over competing approaches~\cite{Gkioxari-CVPR-InteractNet,Gupta-SemanticRoleLabeling,kolesnikov2018detecting}.
%
For HICO-DET, we also demonstrate that our method compares favorably against existing methods~\cite{Gkioxari-CVPR-InteractNet,Chao-WACV-HOI,Shen-WACV-Zeroshot}.
Following the evaluation protocol~\cite{Chao-WACV-HOI}, we report the quantitative evaluation of all, rare, and non-rare interactions with two different settings: `Default' and `Known Object'.
\jiabin{Please update the results here}
Compared to~\cite{Gkioxari-CVPR-InteractNet}, we achieve an absolute gain of $4.90$ points over the best-performing model (InteractNet)~\cite{Gkioxari-CVPR-InteractNet} under the \emph{full category} of the `Default' setting.
This amounts to a relative improvement of $49.3\%$.

\vspace{\secmargin}
\subsection{Qualitative evaluation}
\vspace{\secmargin}

\begin{figure*}[t]
\centering
\footnotesize

\mpage{0.155}{\includegraphics[width=\linewidth]{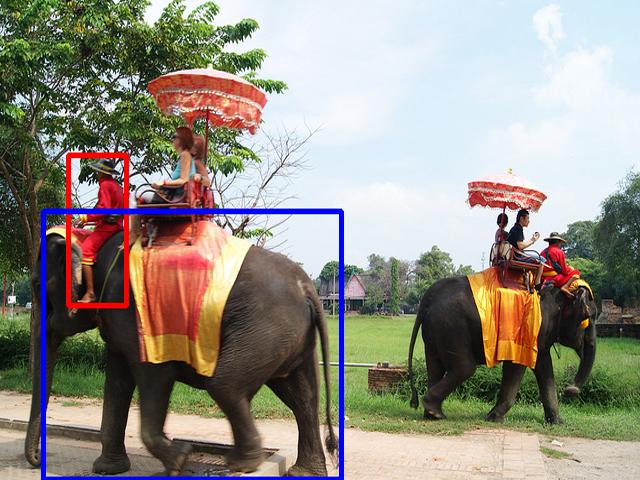}}\hfill
\mpage{0.155}{\includegraphics[width=\linewidth]{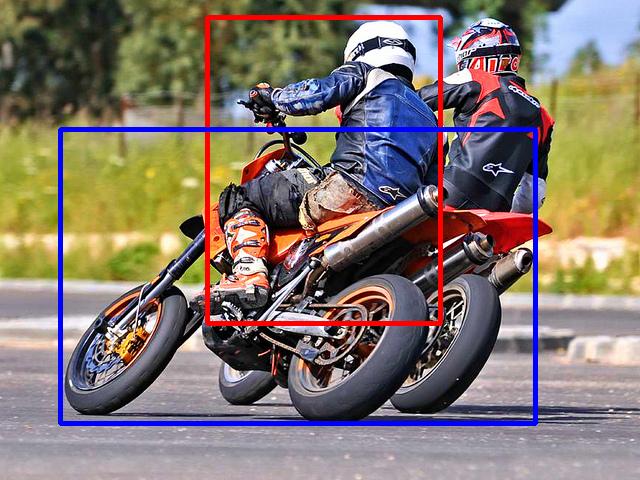}}\hfill
\mpage{0.155}{\includegraphics[width=\linewidth]{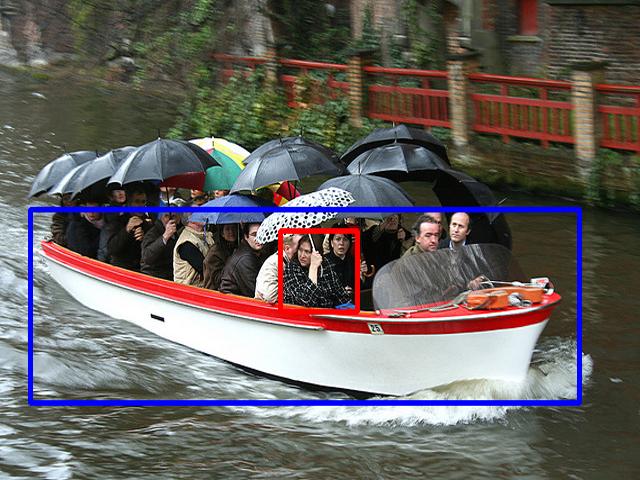}}\hfill
\mpage{0.155}{\includegraphics[width=\linewidth]{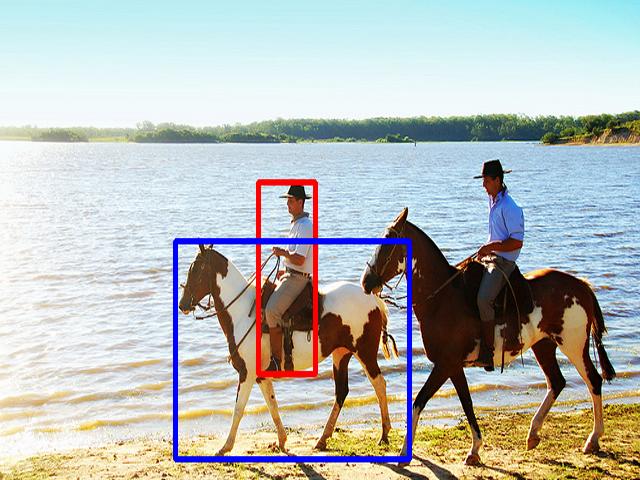}}\hfill
\mpage{0.155}{\includegraphics[width=\linewidth]{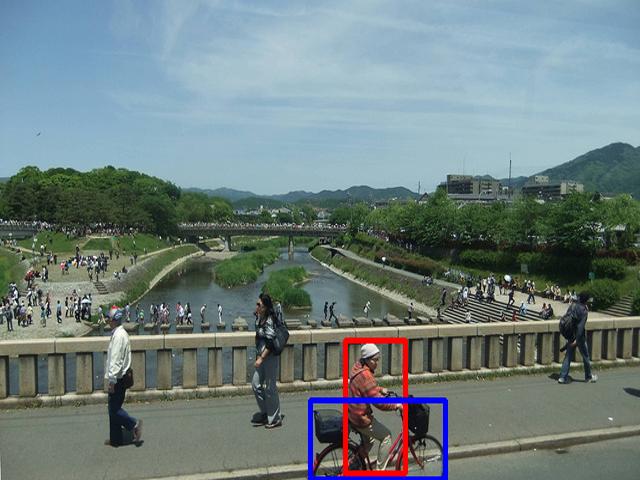}}\hfill
\mpage{0.155}{\includegraphics[width=\linewidth]{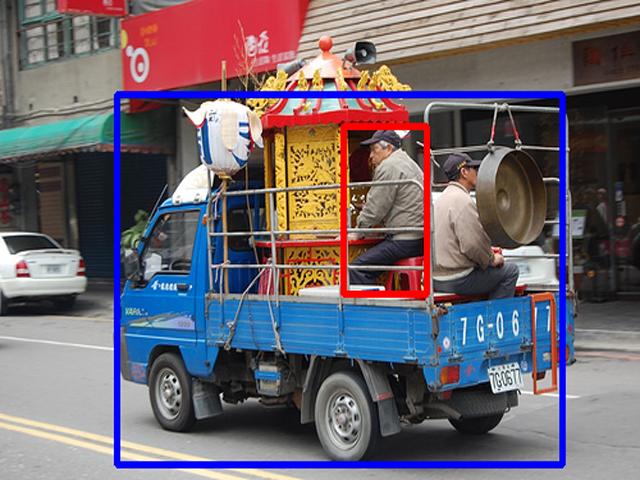}}\hfill
\mpage{0.155}{\scriptsize ride     elephant}\hfill
\mpage{0.155}{\scriptsize ride     motorcycle}\hfill
\mpage{0.155}{\scriptsize ride     boat}\hfill
\mpage{0.155}{\scriptsize ride     horse}\hfill
\mpage{0.155}{\scriptsize ride     bike}\hfill
\mpage{0.155}{\scriptsize ride     truck}\hfill

\mpage{0.155}{\includegraphics[width=\linewidth]{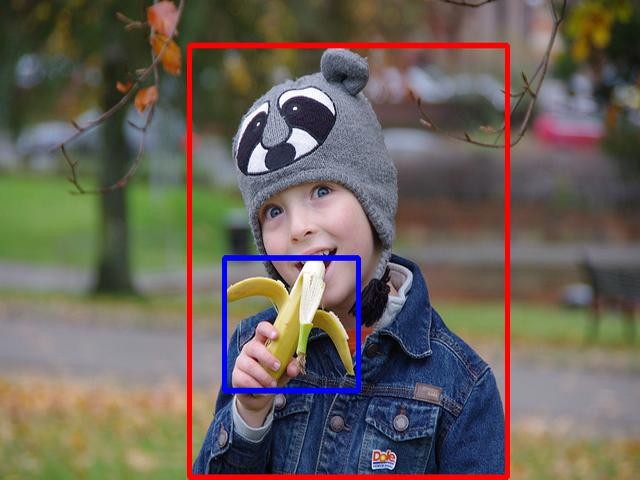}}\hfill
\mpage{0.155}{\includegraphics[width=\linewidth]{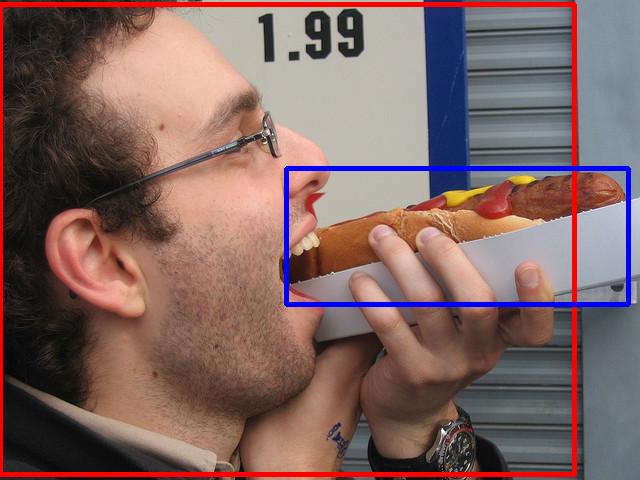}}\hfill
\mpage{0.155}{\includegraphics[width=\linewidth]{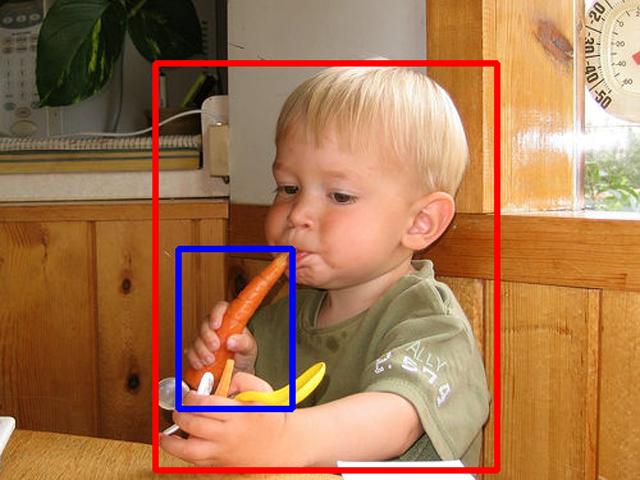}}\hfill
\mpage{0.155}{\includegraphics[width=\linewidth]{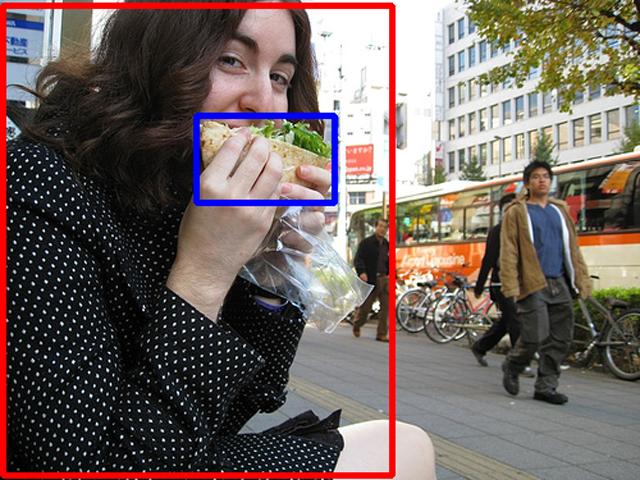}}\hfill
\mpage{0.155}{\includegraphics[width=\linewidth]{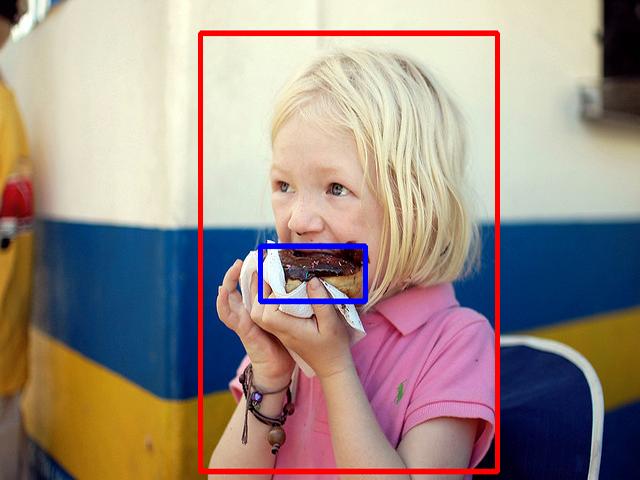}}\hfill
\mpage{0.155}{\includegraphics[width=\linewidth]{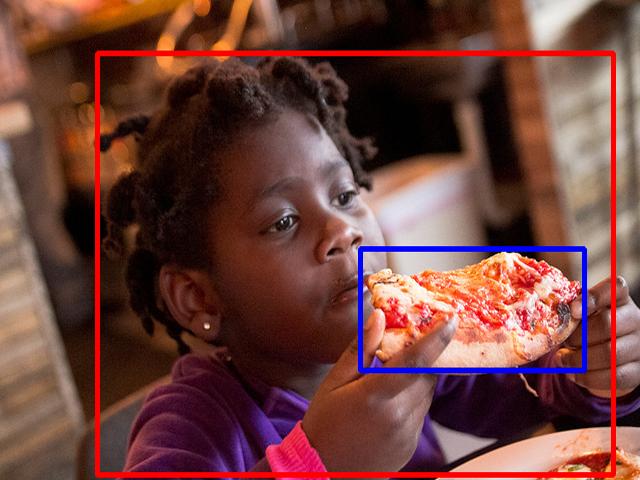}}\hfill
\mpage{0.155}{\scriptsize eat     banana}\hfill
\mpage{0.155}{\scriptsize eat     hot dog}\hfill
\mpage{0.155}{\scriptsize eat     carrot}\hfill
\mpage{0.155}{\scriptsize eat     sandwich}\hfill
\mpage{0.155}{\scriptsize eat     donut}\hfill
\mpage{0.155}{\scriptsize eat     pizza}\hfill

\mpage{0.155}{\includegraphics[width=\linewidth]{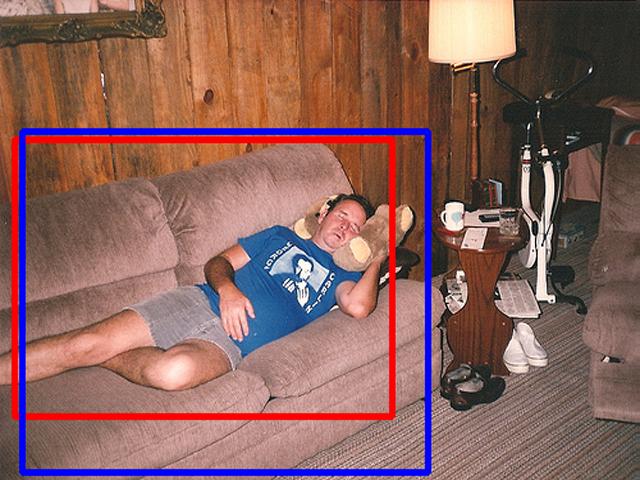}}\hfill
\mpage{0.155}{\includegraphics[width=\linewidth]{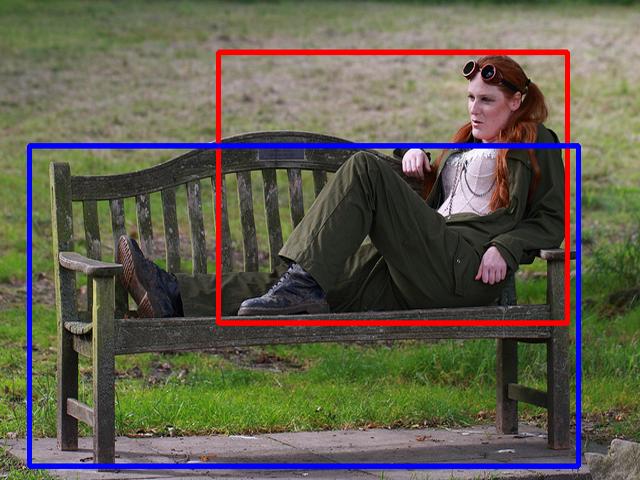}}\hfill
\mpage{0.155}{\includegraphics[width=\linewidth]{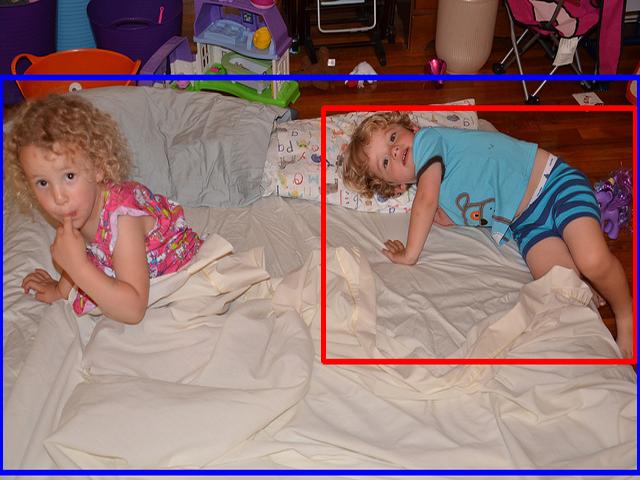}}\hfill
\mpage{0.155}{\includegraphics[width=\linewidth]{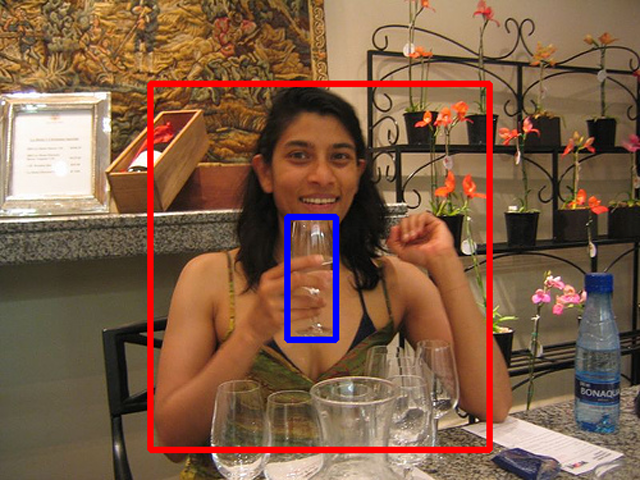}}\hfill
\mpage{0.155}{\includegraphics[width=\linewidth]{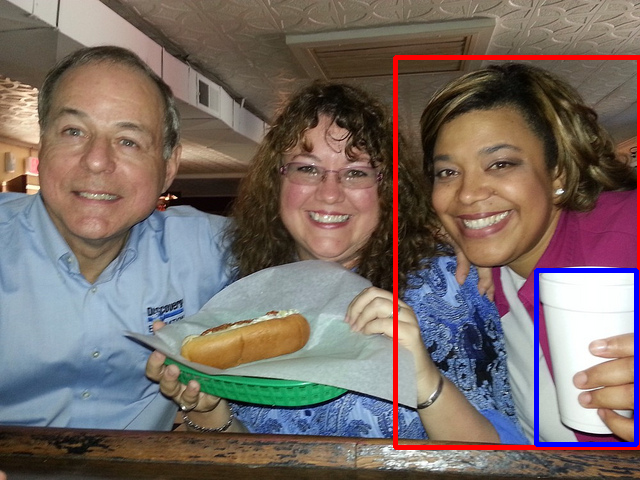}}\hfill
\mpage{0.155}{\includegraphics[width=\linewidth]{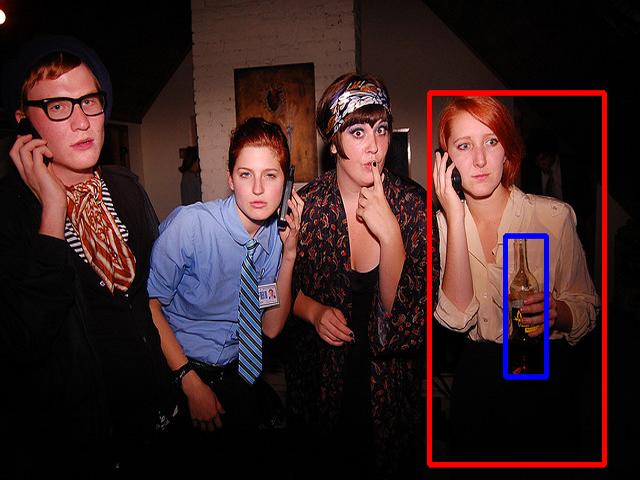}}\hfill
\mpage{0.155}{\scriptsize lay on couch}\hfill
\mpage{0.155}{\scriptsize lay on bench}\hfill
\mpage{0.155}{\scriptsize lay on bed}\hfill
\mpage{0.155}{\scriptsize drink w/ wineglass}\hfill
\mpage{0.155}{\scriptsize drink w/ cup}\hfill
\mpage{0.155}{\scriptsize drink w/ bottle}\hfill
\\
\vspace{\figmargin}
\caption{\textbf{Sample HOI detections on the V-COCO $\emph{test}$ set.} Our model detects various forms of HOIs in everyday photos. For actions `ride', `eat', `lay' and `drink', our model detects a diverse set of objects that the persons are interacting with in different situations. 
\chen{The third row doesn't align well.}
}
\vspace{\figbmargin}
\label{fig:one_action}
\end{figure*}

\begin{figure*}[t]
\centering
\footnotesize

\mpage{0.155}{\includegraphics[width=\linewidth]{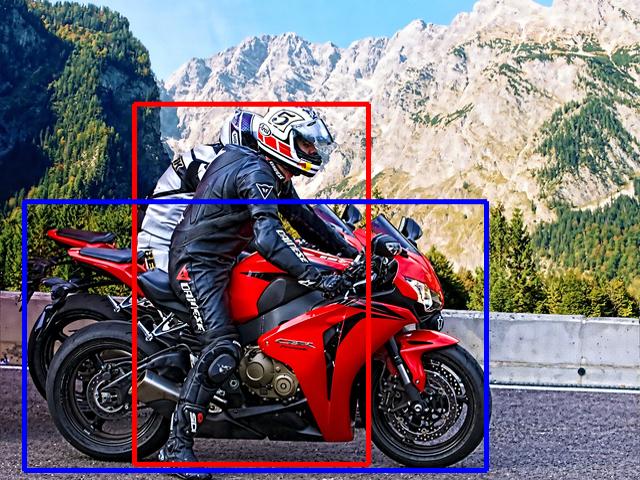}}\hfill
\mpage{0.155}{\includegraphics[width=\linewidth]{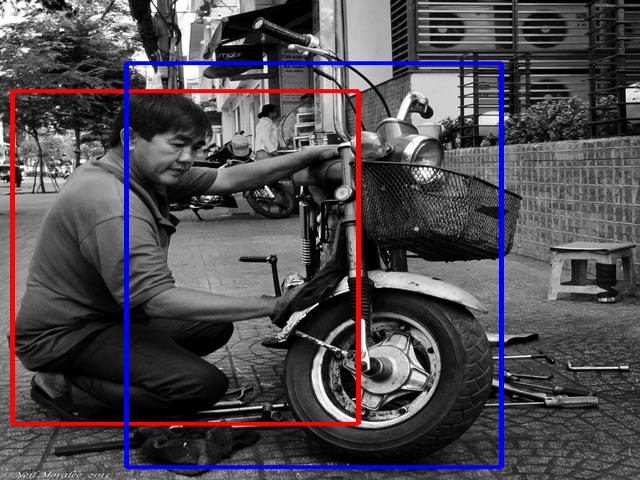}}\hfill
\mpage{0.155}{\includegraphics[width=\linewidth]{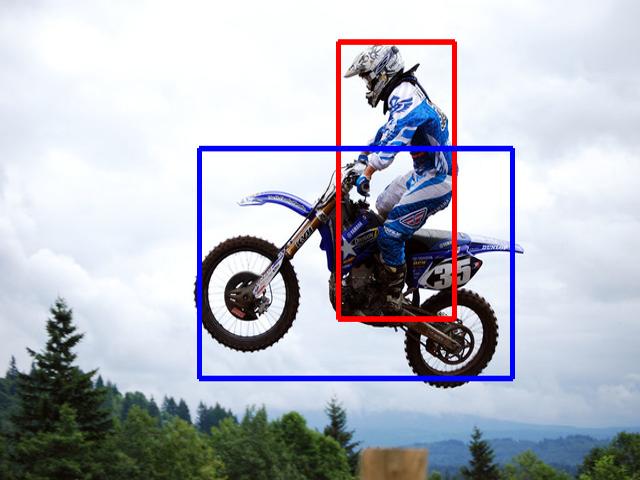}}\hfill
\mpage{0.155}{\includegraphics[width=\linewidth]{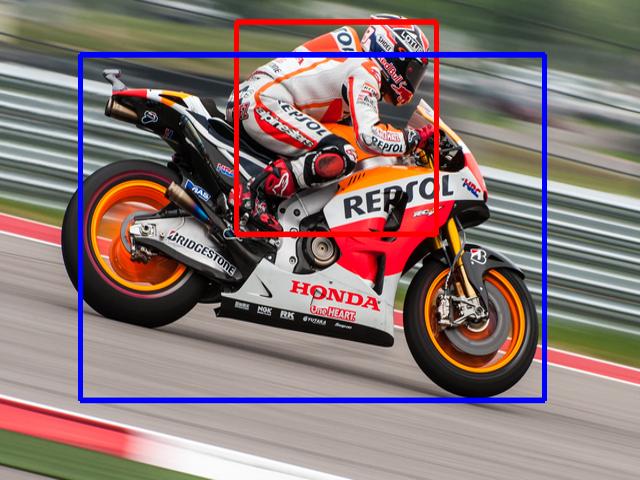}}\hfill
\mpage{0.155}{\includegraphics[width=\linewidth]{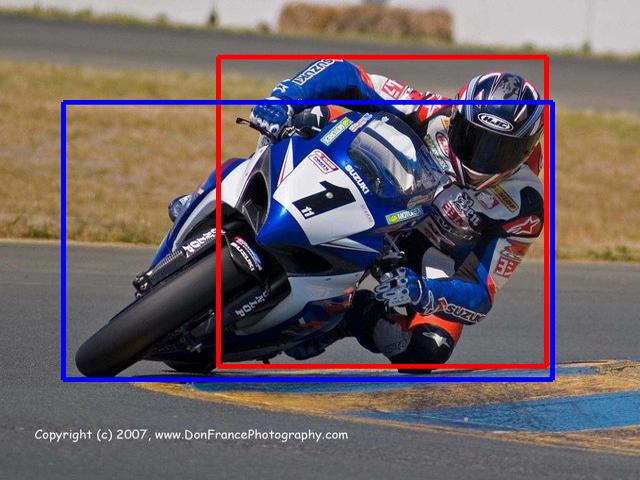}}\hfill
\mpage{0.155}{\includegraphics[width=\linewidth]{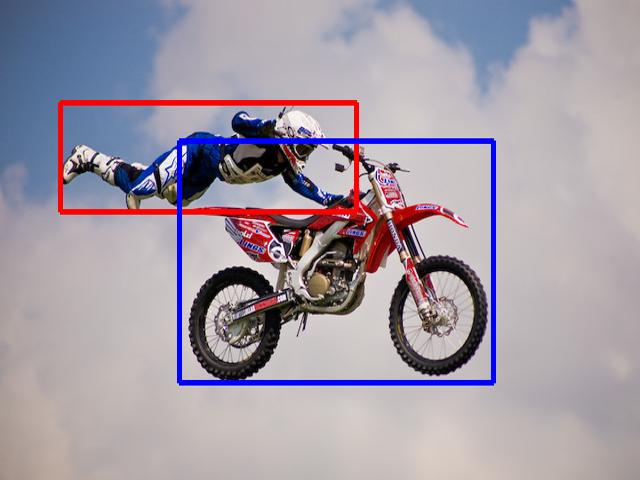}}\hfill

\mpage{0.155}{\scriptsize hold       motorcycle}\hfill
\mpage{0.155}{\scriptsize inspect       motorcycle}\hfill
\mpage{0.155}{\scriptsize jump       motorcycle}\hfill
\mpage{0.155}{\scriptsize race       motorcycle}\hfill
\mpage{0.155}{\scriptsize turn       motorcycle}\hfill
\mpage{0.155}{\scriptsize straddle   motorcycle}\hfill

\mpage{0.155}{\includegraphics[width=\linewidth]{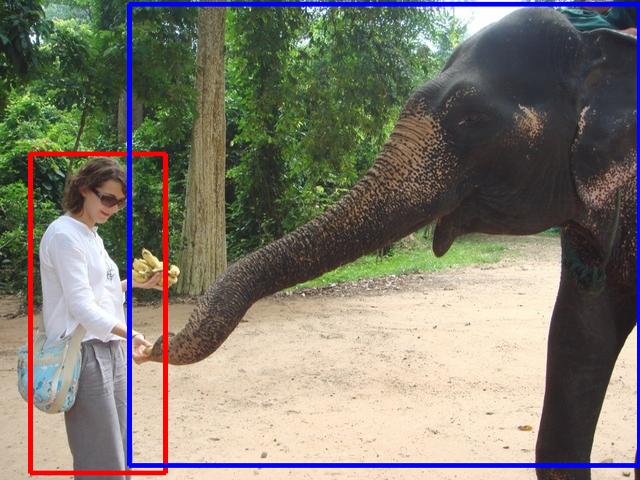}}\hfill
\mpage{0.155}{\includegraphics[width=\linewidth]{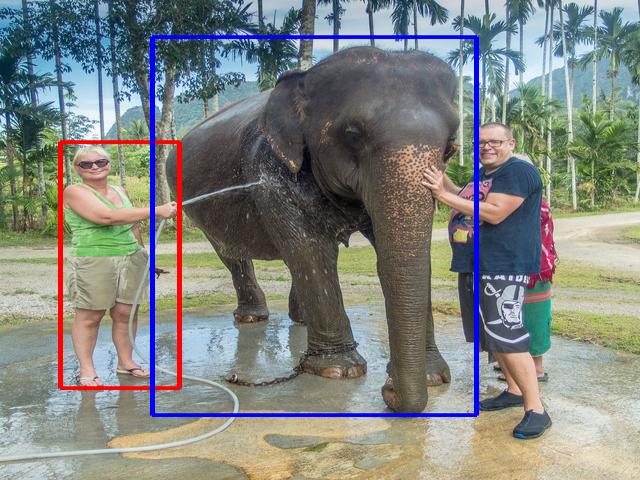}}\hfill
\mpage{0.155}{\includegraphics[width=\linewidth]{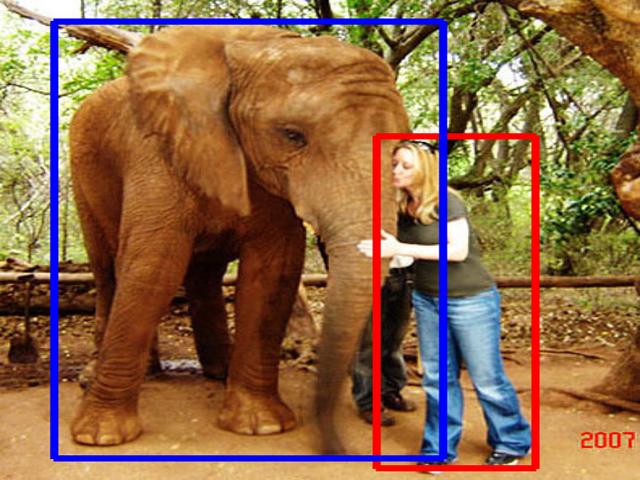}}\hfill
\mpage{0.155}{\includegraphics[width=\linewidth]{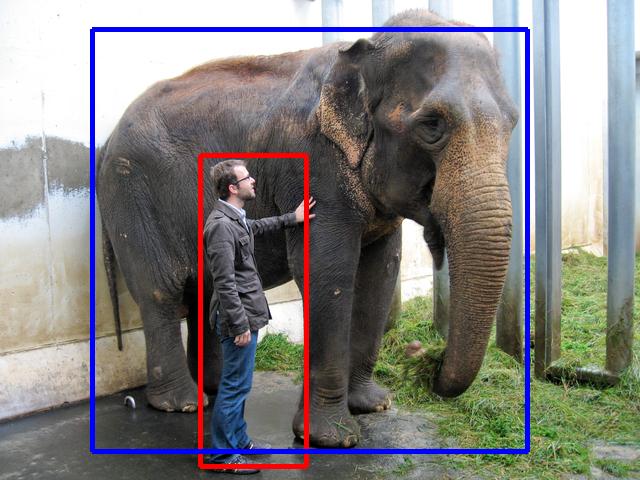}}\hfill
\mpage{0.155}{\includegraphics[width=\linewidth]{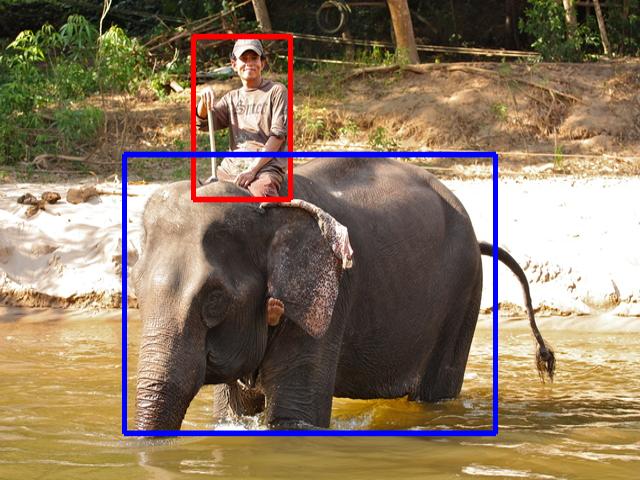}}\hfill
\mpage{0.155}{\includegraphics[width=\linewidth]{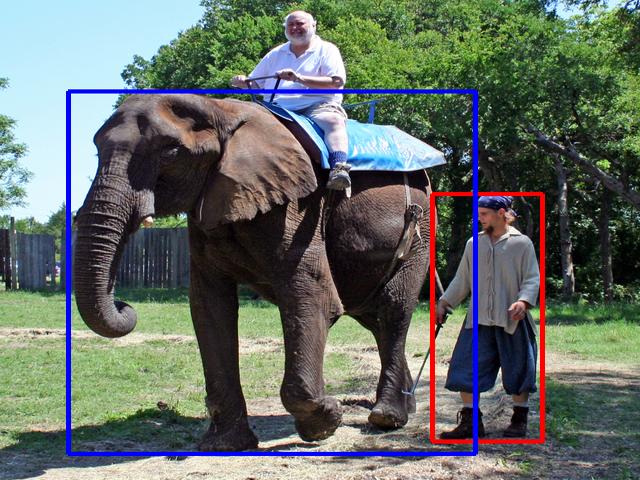}}\hfill

\mpage{0.155}{\scriptsize feed       elephant}\hfill
\mpage{0.155}{\scriptsize hose       elephant}\hfill
\mpage{0.155}{\scriptsize kiss       elephant}\hfill
\mpage{0.155}{\scriptsize pet       elephant}\hfill
\mpage{0.155}{\scriptsize ride       elephant}\hfill
\mpage{0.155}{\scriptsize walk       elephant}\hfill

\vspace{\figmargin}
\caption{\textbf{Sample HOI detections on the HICO-DET $\emph{test}$ set.} Our model detects different types of interactions with objects from the same category.
}
\vspace{\figbmargin}
\label{fig:one_object}
\end{figure*}
\begin{figure*}[t]
\centering


\mpage{0.155}{\includegraphics[width=\linewidth]{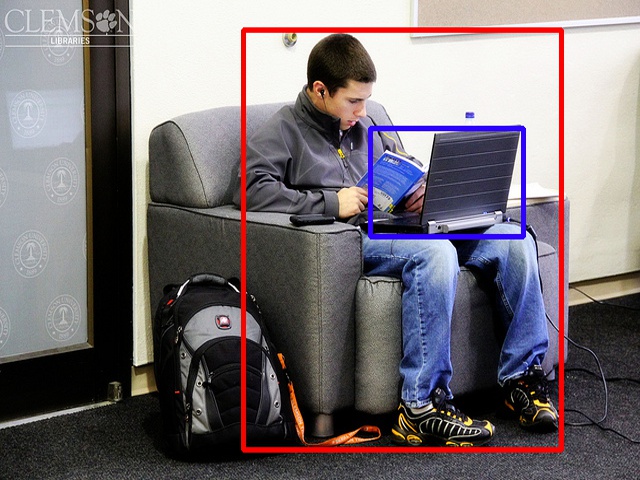}}\hfill
\mpage{0.155}{\includegraphics[width=\linewidth]{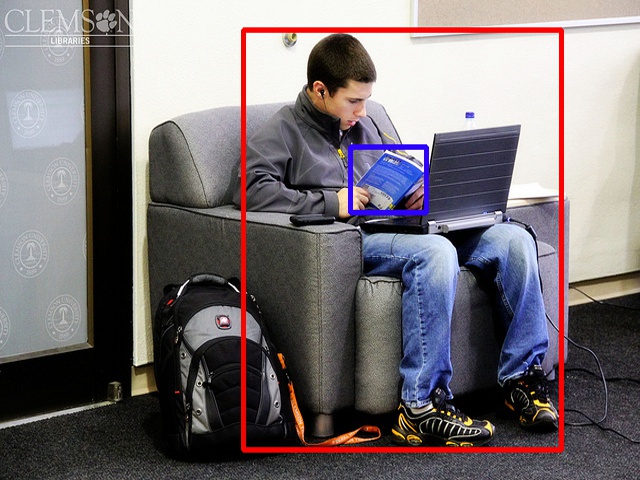}}\hfill
\mpage{0.155}{\includegraphics[width=\linewidth]{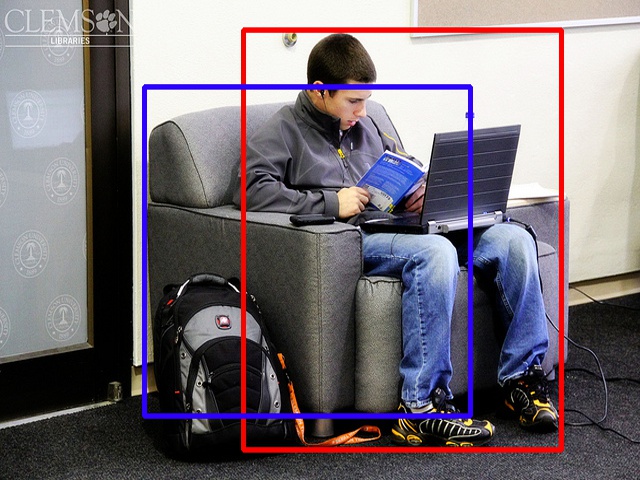}}\hfill
\mpage{0.155}{\includegraphics[width=\linewidth]{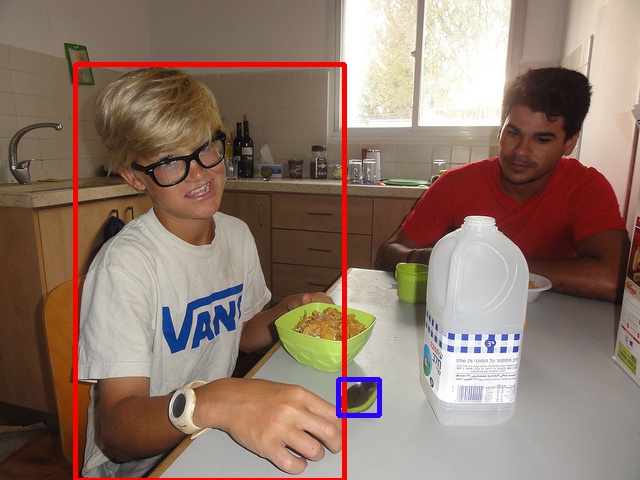}}\hfill
\mpage{0.155}{\includegraphics[width=\linewidth]{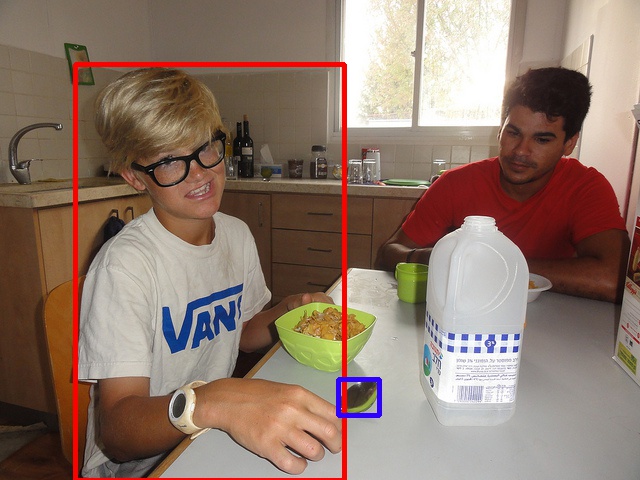}}\hfill
\mpage{0.155}{\includegraphics[width=\linewidth]{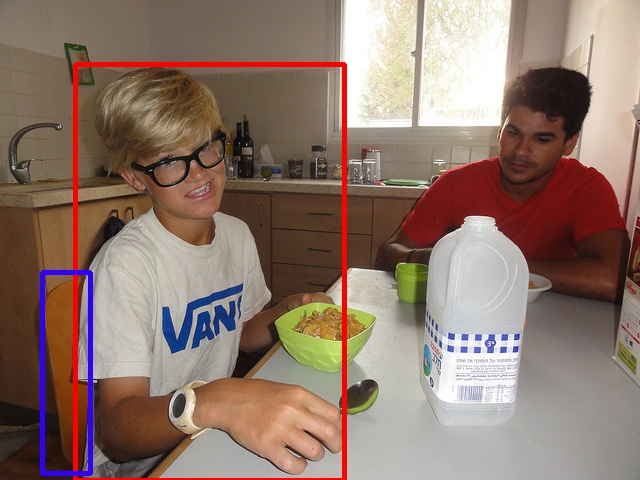}}

\mpage{0.155}{\scriptsize work on laptop}\hfill
\mpage{0.155}{\scriptsize read book}\hfill
\mpage{0.155}{\scriptsize sit on couch}\hfill
\mpage{0.155}{\scriptsize hold spoon}\hfill
\mpage{0.155}{\scriptsize eat with spoon}\hfill
\mpage{0.155}{\scriptsize sit on chair}\\

\vspace{\figmargin}
\caption{\tb{Detecting multiple actions.} Our model detects an individual taking multiple actions and interacting with different objects, \eg the person sitting on the couch is reading a book while working on a laptop.
}
%
%
%
\vspace{\figbmargin}
\label{fig:multi_action}
\end{figure*}

\begin{figure*}[t]
\centering
\footnotesize

\mpage{0.153}{\includegraphics[width=\linewidth]{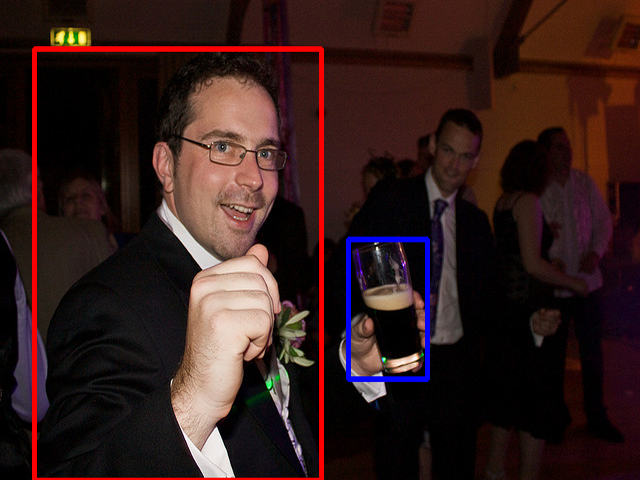}}
\mpage{0.153}{\includegraphics[width=\linewidth]{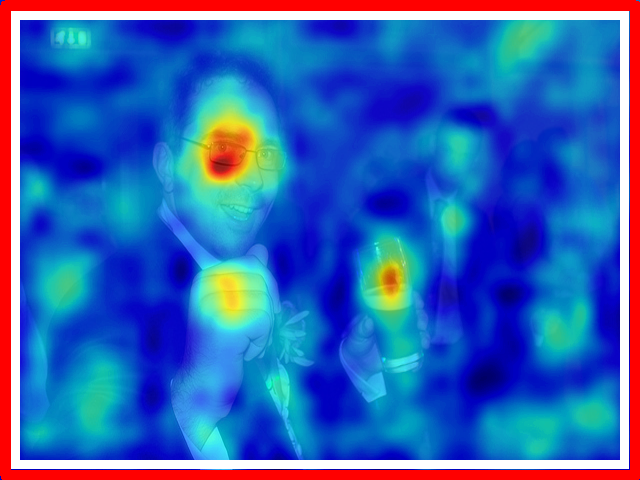}}
\mpage{0.153}{\includegraphics[width=\linewidth]{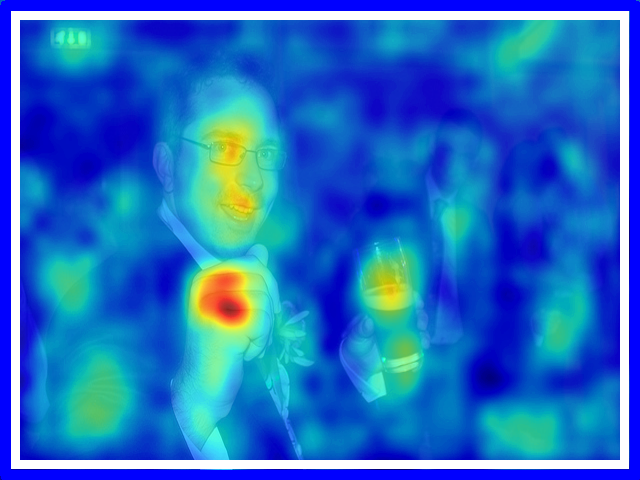}}
\hfill
\mpage{0.153}{\includegraphics[width=\linewidth]{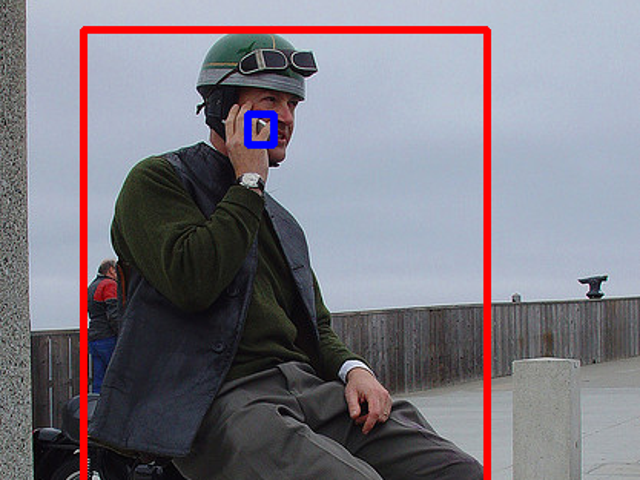}}
\mpage{0.153}{\includegraphics[width=\linewidth]{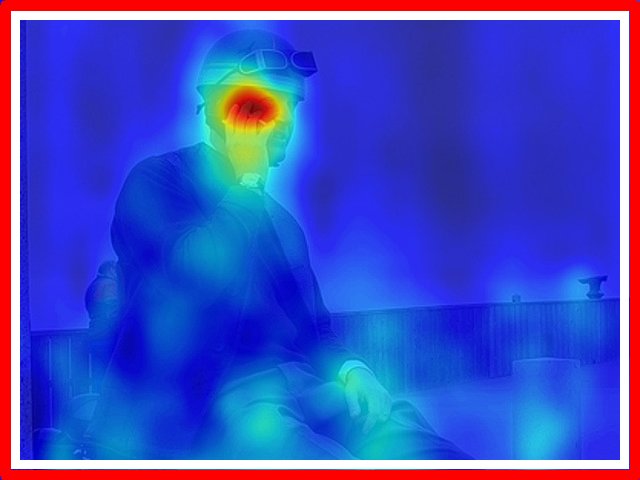}}
\mpage{0.153}{\includegraphics[width=\linewidth]{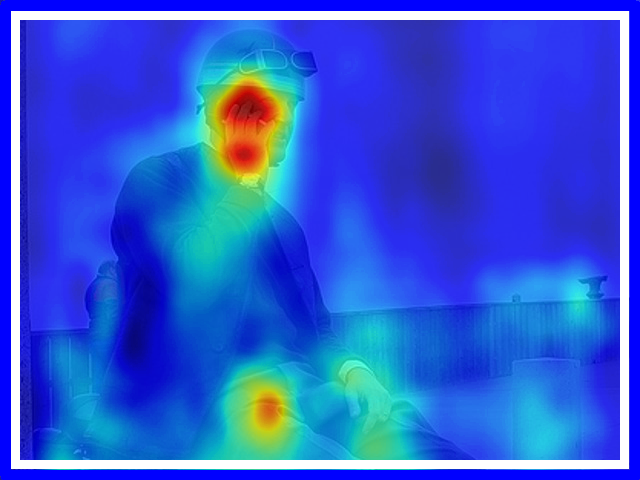}} 

\mpage{0.153}{\scriptsize hold cup}
\mpage{0.153}{\scriptsize Human-centric att.}
\mpage{0.153}{\scriptsize Object-centric att.} 
\hfill
\mpage{0.153}{\scriptsize talk on cellphone}
\mpage{0.153}{\scriptsize Human-centric att.}
\mpage{0.153}{\scriptsize Object-centric att.}

\vspace{1mm}

\mpage{0.22}{\includegraphics[width=\linewidth]{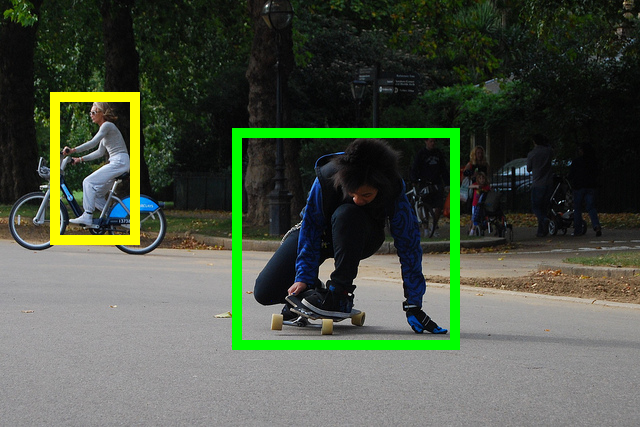}}
\hfill
\mpage{0.15}{\includegraphics[width=\linewidth]{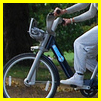}}
\mpage{0.15}{\includegraphics[width=\linewidth]{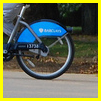}}
\hfill
\mpage{0.15}{\includegraphics[width=\linewidth]{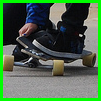}}
\mpage{0.15}{\includegraphics[width=\linewidth]{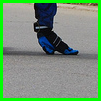}}

\vspace{\figmargin}
\caption{\textbf{Attention map visualization.} (\emph{Top}) Examples of human/object-centric attention maps. (\emph{Bottom}) $100 \times 100$ patches centered at the peaks of the human-centric attentional maps generated by the two persons. Our model learns to attend to objects (e.g., bicycle, skateboard) and the human poses.
}
\vspace{\figbmargin}
\vspace{\figbmargin}
\label{fig:att_HO}
\end{figure*}

\para{HOI detection results.}
Here we show sample HOI detection results on the V-COCO dataset and the HICO-DET dataset. 
We highlight the detected human and object with red and blue bounding boxes, respectively.
\figref{one_action} shows that our model can predict HOIs in a wide variety of different situations.
\figref{one_object} shows that our model is capable of predicting different action with the objects from the same category.
%
\figref{multi_action} presents two examples of detecting a person interacting with different objects. 

\para{Attention map visualization.}
\figref{att_HO} visualizes the human-centric and object-centric attention maps.
The human-centric attention map often focuses on the surrounding objects that help disambiguate action prediction for the detected person.
The object-centric attention map, on the other hand, highlights the informative human body part, \eg in the first image, the attention map highlights the right hand of the person even though he was not holding anything with his right hand.
%
We also show an example of two detected persons doing different actions.
We show the cropped $100 \times 100$ patches centered at the peaks of the generated human-centric attentional maps. 
The highlighted regions roughly correspond to the objects they are interacting with.
%


\vspace{-1mm}
\begin{table}[th]
\caption{{Ablation study on the V-COCO \emph{test} dataset} \jiabin{Need to update all three parts. In particular, the (c) is not consistent with the rest of the results.}\chen{Will do. Waiting for the results.}}
\label{tab:vcoco_scene}
\vspace{0.6mm}
\mpage{0.28}{
\centering
\footnotesize{
\begin{tabular}{l|c}
\toprule
 & $AP_{role}$ \\
\midrule
None & 42.5 \\
Full image ~\cite{Mallya-ECCV-Interactions} & 42.9 \\
Bottom-up att.~\cite{Girdhar-NIPS-AttentionalPooling} 
& $\underline{43.2}$ \\
Inst-centric att. (ours)                                & \tb{44.7} \\
\bottomrule
\end{tabular}
}
}
\hfill
\mpage{0.28}{ \centering
\footnotesize{
\begin{tabular}{cc|c}
\toprule
   Human   &    Object   & $AP_{role}$ \\
\midrule
     -     &      -     &  42.5 \\
\checkmark &      -     &  $\underline{44.4}$ \\
     -     & \checkmark &  44.3 \\
\checkmark & \checkmark & \tb{44.7} \\
\bottomrule
\end{tabular}
}
}
\hfill
\mpage{0.35}{
\vspace{0.5mm}
\includegraphics[width=\linewidth]{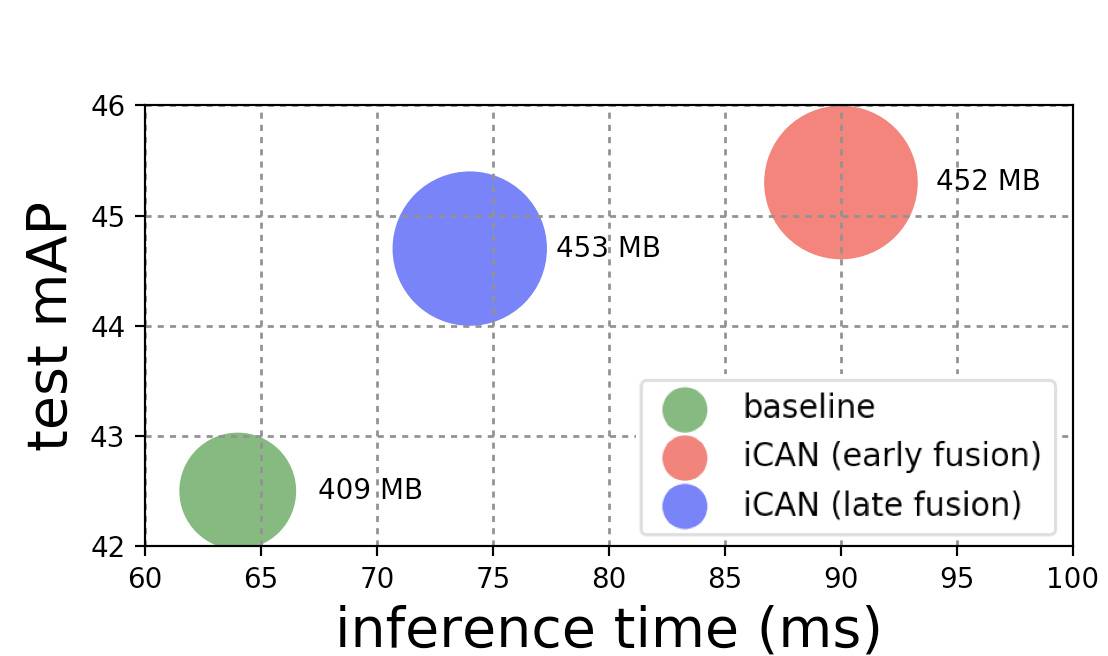}
}
\\
\vspace{-5mm}\\
\mpage{0.28}{(a) Scene feature} 
\hfill
\mpage{0.28}{(b) Human/Object stream} 
\hfill
\mpage{0.34}{(c) mAP \vs time/model size}
\vspace{\tabmargin}
\end{table}

\vspace{\secmargin}
\subsection{Ablation study}
\vspace{\secmargin}


\para{Contextual feature.} Recognizing the correct actions using only human and object appearance features remains challenging.
Building upon a strong baseline that does not use any contextual features, we investigate several different approaches for incorporating contextual information of the image, including bottom-up attention map~\cite{Girdhar-NIPS-AttentionalPooling}, convolutional features from the entire image~\cite{Mallya-ECCV-Interactions}, and the proposed instance-centric attention map.
\tabref{vcoco_scene}(a) shows that incorporating contextual features generally helps improve the HOI detection performance.
%
%
Our approach provides a larger boost over methods that use features without conditioning on the instance-of-interest.

%
%

\para{Human-centric \vs object-centric.} \tabref{vcoco_scene}(b) validates the importance of leveraging both human-centric and object-centric attentional maps.
%

\para{mAP \vs time \vs memory.} \tabref{vcoco_scene}(c) characterizes the variants of the proposed iCAN using their trade-off in terms of mAP, inference time, and memory usage.
Our model with early fusion achieves the best performance on V-COCO dataset. 
However, this comes at the cost of expensive evaluation of all possible human-object pairs in an image based on their appearance features, and slower training and testing time.

\vspace{-4mm}
\section{Conclusions}
\label{sec:conclusions}
\vspace{\secmargin}
In this paper, we propose an instance-centric attention module for HOI detection.
Our core idea is to learn to highlight informative regions from an image using the appearance of a person and an object instance, which allow us to gather relevant contextual information facilitating HOI detection.
We validate the effectiveness of our approach and show a sizable performance boost compared to the state-of-the-arts on two HOI benchmark datasets.
%
%
In this work we consider class-agnostic instance-centric attention. 
We believe that the class-dependent instance-centric attention is a promising future direction.

\textbf{Acknowledgements.} This work was supported in part by NSF under Grant No. (1755785). We gratefully acknowledge the support of NVIDIA Corporation with the donation of the Titan Xp GPU.

\clearpage
{\small
\bibliographystyle{splncs}
\bibliography{main}

\begin{thebibliography}{44}
\providecommand{\natexlab}[1]{#1}
\providecommand{\url}[1]{\texttt{#1}}
\expandafter\ifx\csname urlstyle\endcsname\relax
  \providecommand{\doi}[1]{doi: #1}\else
  \providecommand{\doi}{doi: \begingroup \urlstyle{rm}\Url}\fi

\bibitem[Bilen and Vedaldi(2016)]{Bilen-CVPR-Weakly}
Hakan Bilen and Andrea Vedaldi.
\newblock Weakly supervised deep detection networks.
\newblock In \emph{CVPR}, 2016.

\bibitem[Cao et~al.(2017)Cao, Simon, Wei, and Sheikh]{Cao-CVPR-OpenPese}
Zhe Cao, Tomas Simon, Shih-En Wei, and Yaser Sheikh.
\newblock Realtime multi-person 2d pose estimation using part affinity fields.
\newblock In \emph{CVPR}, 2017.

\bibitem[Chao et~al.(2015)Chao, Wang, He, Wang, and Deng]{Chao-CVPR-HICO}
Yu-Wei Chao, Zhan Wang, Yugeng He, Jiaxuan Wang, and Jia Deng.
\newblock {HICO}: A benchmark for recognizing human-object interactions in
  images.
\newblock In \emph{CVPR}, 2015.

\bibitem[Chao et~al.(2017)Chao, Liu, Liu, Zeng, and Deng]{Chao-WACV-HOI}
Yu-Wei Chao, Yunfan Liu, Xieyang Liu, Huayi Zeng, and Jia Deng.
\newblock Learning to detect human-object interactions.
\newblock In \emph{WACV}, 2017.

\bibitem[Chen et~al.(2017)Chen, Papandreou, Kokkinos, Murphy, and
  Yuille]{Chen-TPAMI-DeepLab}
Liang-Chieh Chen, George Papandreou, Iasonas Kokkinos, Kevin Murphy, and Alan~L
  Yuille.
\newblock {Deep{L}ab}: Semantic image segmentation with deep convolutional
  nets, atrous convolution, and fully connected crfs.
\newblock \emph{TPAMI}, 40\penalty0 (4), 2017.

\bibitem[Ch{\'e}ron et~al.(2015)Ch{\'e}ron, Laptev, and
  Schmid]{Cheron-ICCV-PCNN}
Guilhem Ch{\'e}ron, Ivan Laptev, and Cordelia Schmid.
\newblock {P-CNN}: Pose-based cnn features for action recognition.
\newblock In \emph{ICCV}, 2015.

\bibitem[Dai et~al.(2017)Dai, Zhang, and Lin]{Dai-CVPR-Relationship}
Bo~Dai, Yuqi Zhang, and Dahua Lin.
\newblock Detecting visual relationships with deep relational networks.
\newblock In \emph{CVPR}, 2017.

\bibitem[Dai et~al.(2016)Dai, Li, He, and Sun]{Dai-NIPS-RFCN}
Jifeng Dai, Yi~Li, Kaiming He, and Jian Sun.
\newblock {R-FCN}: Object detection via region-based fully convolutional
  networks.
\newblock In \emph{NIPS}, 2016.

\bibitem[Girdhar and Ramanan(2017)]{Girdhar-NIPS-AttentionalPooling}
Rohit Girdhar and Deva Ramanan.
\newblock Attentional pooling for action recognition.
\newblock In \emph{NIPS}, 2017.

\bibitem[Girshick(2015)]{Ross-CVPR-FastRCNN}
Ross Girshick.
\newblock Fast r-cnn.
\newblock In \emph{CVPR}, 2015.

\bibitem[Girshick et~al.(2014)Girshick, Donahue, Darrell, and
  Malik]{Ross-CVPR-Hierarchies}
Ross Girshick, Jeff Donahue, Trevor Darrell, and Jitendra Malik.
\newblock Rich feature hierarchies for accurate object detection and semantic
  segmentation.
\newblock In \emph{CVPR}, 2014.

\bibitem[Girshick et~al.(2018)Girshick, Radosavovic, Gkioxari, Doll\'{a}r, and
  He]{Detectron}
Ross Girshick, Ilija Radosavovic, Georgia Gkioxari, Piotr Doll\'{a}r, and
  Kaiming He.
\newblock Detectron.
\newblock \url{https://github.com/facebookresearch/detectron}, 2018.

\bibitem[Gkioxari et~al.(2015)Gkioxari, Girshick, and
  Malik]{Gkioxari-ICCV-R*CNN}
Georgia Gkioxari, Ross Girshick, and Jitendra Malik.
\newblock Contextual action recognition with r* cnn.
\newblock In \emph{ICCV}, 2015.

\bibitem[Gkioxari et~al.(2018)Gkioxari, Girshick, Doll{\'a}r, and
  He]{Gkioxari-CVPR-InteractNet}
Georgia Gkioxari, Ross Girshick, Piotr Doll{\'a}r, and Kaiming He.
\newblock Detecting and recognizing human-object interactions.
\newblock In \emph{CVPR}, 2018.

\bibitem[Gupta et~al.(2009)Gupta, Kembhavi, and Davis]{Gupta-TPAMI-Observing}
Abhinav Gupta, Aniruddha Kembhavi, and Larry~S Davis.
\newblock Observing human-object interactions: Using spatial and functional
  compatibility for recognition.
\newblock \emph{TPAMI}, 31\penalty0 (10), 2009.

\bibitem[Gupta and Malik(2015)]{Gupta-SemanticRoleLabeling}
Saurabh Gupta and Jitendra Malik.
\newblock Visual semantic role labeling.
\newblock \emph{arXiv preprint arXiv:1505.04474}, 2015.

\bibitem[He et~al.(2016)He, Zhang, Ren, and Sun]{He-CVPR-ResNet}
Kaiming He, Xiangyu Zhang, Shaoqing Ren, and Jian Sun.
\newblock Deep residual learning for image recognition.
\newblock In \emph{CVPR}, 2016.

\bibitem[He et~al.(2017)He, Gkioxari, Doll{\'a}r, and
  Girshick]{He-ICCV-MaskRCNN}
Kaiming He, Georgia Gkioxari, Piotr Doll{\'a}r, and Ross Girshick.
\newblock Mask r-cnn.
\newblock In \emph{ICCV}, 2017.

\bibitem[Hu et~al.(2017)Hu, Rohrbach, Andreas, Darrell, and
  Saenko]{Hu-CVPR-Referential}
Ronghang Hu, Marcus Rohrbach, Jacob Andreas, Trevor Darrell, and Kate Saenko.
\newblock Modeling relationships in referential expressions with compositional
  modular networks.
\newblock In \emph{CVPR}, 2017.

\bibitem[Jetley et~al.(2018)Jetley, Lord, Lee, and
  Torr]{Jetley-ICLR-PayAttention}
Saumya Jetley, Nicholas~A Lord, Namhoon Lee, and Philip~HS Torr.
\newblock Learn to pay attention.
\newblock In \emph{ICLR}, 2018.

\bibitem[Johnson et~al.(2015)Johnson, Krishna, Stark, Li, Shamma, Bernstein,
  and Fei-Fei]{Johnson-CVPR-Retrieval}
Justin Johnson, Ranjay Krishna, Michael Stark, Li-Jia Li, David Shamma, Michael
  Bernstein, and Li~Fei-Fei.
\newblock Image retrieval using scene graphs.
\newblock In \emph{CVPR}, 2015.

\bibitem[Kolesnikov et~al.(2018)Kolesnikov, Lampert, and
  Ferrari]{kolesnikov2018detecting}
Alexander Kolesnikov, Christoph~H Lampert, and Vittorio Ferrari.
\newblock Detecting visual relationships using box attention.
\newblock \emph{arXiv preprint arXiv:1807.02136}, 2018.

\bibitem[Li et~al.(2017{\natexlab{a}})Li, Ouyang, Wang, and Tang]{Li-CVPR-VIP}
Yikang Li, Wanli Ouyang, Xiaogang Wang, and Xiao'ou Tang.
\newblock {ViP-CNN}: Visual phrase guided convolutional neural network.
\newblock In \emph{CVPR}, 2017{\natexlab{a}}.

\bibitem[Li et~al.(2017{\natexlab{b}})Li, Ouyang, Zhou, Wang, and
  Wang]{Li-CVPR-Scene}
Yikang Li, Wanli Ouyang, Bolei Zhou, Kun Wang, and Xiaogang Wang.
\newblock Scene graph generation from objects, phrases and region captions.
\newblock In \emph{CVPR}, 2017{\natexlab{b}}.

\bibitem[Li et~al.(2017{\natexlab{c}})Li, Ouyang, Zhou, Wang, and
  Wang]{li2017scene}
Yikang Li, Wanli Ouyang, Bolei Zhou, Kun Wang, and Xiaogang Wang.
\newblock Scene graph generation from objects, phrases and region captions.
\newblock In \emph{ICCV}, 2017{\natexlab{c}}.

\bibitem[Lin et~al.(2014)Lin, Maire, Belongie, Hays, Perona, Ramanan,
  Doll{\'a}r, and Zitnick]{Lin-ECCV-MSCOCO}
Tsung-Yi Lin, Michael Maire, Serge Belongie, James Hays, Pietro Perona, Deva
  Ramanan, Piotr Doll{\'a}r, and C~Lawrence Zitnick.
\newblock Microsoft {COCO}: Common objects in context.
\newblock In \emph{ECCV}, 2014.

\bibitem[Lin et~al.(2017)Lin, Doll{\'a}r, Girshick, He, Hariharan, and
  Belongie]{Lin-CVPR-Pyramid}
Tsung-Yi Lin, Piotr Doll{\'a}r, Ross Girshick, Kaiming He, Bharath Hariharan,
  and Serge Belongie.
\newblock Feature pyramid networks for object detection.
\newblock In \emph{CVPR}, 2017.

\bibitem[Long et~al.(2015)Long, Shelhamer, and Darrell]{Long-CVPR-FCN}
Jonathan Long, Evan Shelhamer, and Trevor Darrell.
\newblock Fully convolutional networks for semantic segmentation.
\newblock In \emph{CVPR}, 2015.

\bibitem[Lu et~al.(2016)Lu, Krishna, Bernstein, and Fei-Fei]{Lu-ECCV-Prior}
Cewu Lu, Ranjay Krishna, Michael Bernstein, and Li~Fei-Fei.
\newblock Visual relationship detection with language priors.
\newblock In \emph{ECCV}, 2016.

\bibitem[Maji et~al.(2011)Maji, Bourdev, and Malik]{Maji-CVPR-Action}
Subhransu Maji, Lubomir Bourdev, and Jitendra Malik.
\newblock Action recognition from a distributed representation of pose and
  appearance.
\newblock In \emph{CVPR}, 2011.

\bibitem[Mallya and Lazebnik(2016)]{Mallya-ECCV-Interactions}
Arun Mallya and Svetlana Lazebnik.
\newblock Learning models for actions and person-object interactions with
  transfer to question answering.
\newblock In \emph{ECCV}, 2016.

\bibitem[Peyre et~al.(2017)Peyre, Laptev, Schmid, and Sivic]{Peyre-ICCV-Weakly}
Julia Peyre, Ivan Laptev, Cordelia Schmid, and Josef Sivic.
\newblock Weakly-supervised learning of visual relations.
\newblock In \emph{ICCV}, 2017.

\bibitem[Plummer et~al.(2017)Plummer, Mallya, Cervantes, Hockenmaier, and
  Lazebnik]{Plummer-ICCV-Phrase}
Bryan~A Plummer, Arun Mallya, Christopher~M Cervantes, Julia Hockenmaier, and
  Svetlana Lazebnik.
\newblock Phrase localization and visual relationship detection with
  comprehensive linguistic cues.
\newblock In \emph{ICCV}, 2017.

\bibitem[Ren et~al.(2015)Ren, He, Girshick, and Sun]{Ren-NIPS-FasterRCNN}
Shaoqing Ren, Kaiming He, Ross Girshick, and Jian Sun.
\newblock {Faster R-CNN}: Towards real-time object detection with region
  proposal networks.
\newblock In \emph{NIPS}, 2015.

\bibitem[Shen et~al.(2018)Shen, Yeung, Hoffman, Mori, and
  Fei-Fei]{Shen-WACV-Zeroshot}
Liyue Shen, Serena Yeung, Judy Hoffman, Greg Mori, and Li~Fei-Fei.
\newblock Scaling human-object interaction recognition through zero-shot
  learning.
\newblock In \emph{WACV}, 2018.

\bibitem[Szegedy et~al.(2016)Szegedy, Vanhoucke, Ioffe, Shlens, and
  Wojna]{szegedy2016rethinking}
Christian Szegedy, Vincent Vanhoucke, Sergey Ioffe, Jon Shlens, and Zbigniew
  Wojna.
\newblock Rethinking the inception architecture for computer vision.
\newblock In \emph{CVPR}, 2016.

\bibitem[Vaswani et~al.(2017)Vaswani, Shazeer, Parmar, Uszkoreit, Jones, Gomez,
  Kaiser, and Polosukhin]{vaswani2017attention}
Ashish Vaswani, Noam Shazeer, Niki Parmar, Jakob Uszkoreit, Llion Jones,
  Aidan~N Gomez, {\L}ukasz Kaiser, and Illia Polosukhin.
\newblock Attention is all you need.
\newblock In \emph{NIPS}, 2017.

\bibitem[Wang et~al.(2018)Wang, Girshick, Gupta, and He]{wang2017non}
Xiaolong Wang, Ross Girshick, Abhinav Gupta, and Kaiming He.
\newblock Non-local neural networks.
\newblock In \emph{CVPR}, 2018.

\bibitem[Xu et~al.(2017)Xu, Zhu, Choy, and Fei-Fei]{Xu-CVPR-SceneGraph}
Danfei Xu, Yuke Zhu, Christopher~B Choy, and Li~Fei-Fei.
\newblock Scene graph generation by iterative message passing.
\newblock In \emph{CVPR}, 2017.

\bibitem[Yao et~al.(2011{\natexlab{a}})Yao, Gall, Fanelli, and
  Van~Gool]{Yao-BMVC-Pose}
Angela Yao, Juergen Gall, Gabriele Fanelli, and Luc Van~Gool.
\newblock Does human action recognition benefit from pose estimation?
\newblock In \emph{BMVC}, 2011{\natexlab{a}}.

\bibitem[Yao et~al.(2011{\natexlab{b}})Yao, Khosla, and
  Fei-Fei]{Yao-CVPR-Combining}
Bangpeng Yao, Aditya Khosla, and Li~Fei-Fei.
\newblock Combining randomization and discrimination for fine-grained image
  categorization.
\newblock In \emph{CVPR}, 2011{\natexlab{b}}.

\bibitem[Zellers et~al.(2018)Zellers, Yatskar, Thomson, and
  Choi]{zellers2018neural}
Rowan Zellers, Mark Yatskar, Sam Thomson, and Yejin Choi.
\newblock {Neural Motifs}: Scene graph parsing with global context.
\newblock In \emph{CVPR}, 2018.

\bibitem[Zhang et~al.(2017)Zhang, Kyaw, Yu, and Chang]{Zhang-ICCV-PPR}
Hanwang Zhang, Zawlin Kyaw, Jinyang Yu, and Shih-Fu Chang.
\newblock {PPR-FCN}: Weakly supervised visual relation detection via parallel
  pairwise r-fcn.
\newblock In \emph{ICCV}, 2017.

\bibitem[Zhuang et~al.(2017)Zhuang, Liu, Shen, and
  Reid]{Zhuang-ICCV-ContextAware}
Bohan Zhuang, Lingqiao Liu, Chunhua Shen, and Ian Reid.
\newblock Towards context-aware interaction recognition for visual relationship
  detection.
\newblock In \emph{ICCV}, 2017.

\end{thebibliography}
}


\end{document}